\pdfoutput=1
\documentclass[11pt]{article}

\usepackage[final]{emnlp2022}

\usepackage{times}
\usepackage{latexsym}

\usepackage[T1]{fontenc}
\usepackage[utf8]{inputenc}

\usepackage{microtype}

\usepackage{helvet} % DO NOT CHANGE THIS
\usepackage{courier}  % DO NOT CHANGE THIS
\usepackage{url}  % DO NOT CHANGE THIS
\usepackage{graphicx} % DO NOT CHANGE THIS

\usepackage[utf8]{inputenc}
\usepackage{subfigure}

\usepackage{amsmath}
\usepackage{amsfonts}
\usepackage{multirow}
\usepackage{color}
\usepackage[lined,boxed,commentsnumbered, ruled,linesnumbered,noend]{algorithm2e}
\usepackage{enumitem}

\usepackage{colortbl}
\usepackage{arydshln}
\usepackage{multicol}
\usepackage[misc]{ifsym}
\usepackage{amssymb}
\usepackage{booktabs}

\title{P$^3$LM: Probabilistically Permuted Prophet Language Modeling \\for Generative Pre-Training}

\author{Junwei Bao\textsuperscript{\dag}\thanks{~~Corresponding author: baojunwei001@gmail.com}, Yifan Wang\textsuperscript{\dag}, Jiangyong Ying\textsuperscript{\ddag}, Yeyun Gong\textsuperscript{$\sharp$},  \\
\bf Jing Zhao\textsuperscript{\dag}, Youzheng Wu\textsuperscript{\dag}, Xiaodong He\textsuperscript{\dag}\\
  \textsuperscript{\dag}JD AI Research~~~~~~~~~~~~~~~~
  \textsuperscript{\ddag}Huawei Technologies~~~~~~~~~~~~~~~~~~
  \textsuperscript{$\sharp$}Microsoft Research Asia\\
  \tt{baojunwei@jd.com yingjiangyong@huawei.com yegong@microsoft.com}\\
  }
  
\date{}
\begin{document}
\maketitle
\begin{abstract}
Conventional autoregressive left-to-right (L2R) sequence generation faces two issues during decoding: limited to unidirectional target sequence modeling, and constrained on strong local dependencies.
To address the aforementioned problem, 
% we study transformer-based language modeling with permuted ordered target sequence and observe that \textit{order matters for sequence generation}.
% Based on this observation, 
we propose \textbf{P$^3$LM}, a \textbf{p}robabilistically \textbf{p}ermuted \textbf{p}rophet \textbf{l}anguage \textbf{m}odel, which strengthens the modeling of bidirectional information and long token dependencies for sequence generation.
Specifically, P$^3$LM learns to generate tokens in permuted order upon an order-aware transformer decoder, as well as to generate the corresponding future $N$ tokens with a multi-stream attention mechanism.
Extensive experiments are conducted on the GLGE benchmark, which includes four datasets for summarization, two for question generation, one for conversational question answering, and one for dialog response generation, where P$^3$LM achieves state-of-the-art results compared with strong publicly available generative pre-training methods.\footnote{The code is available at \url{https://github.com/JunweiBao/P3LM}.}
% \footnote{Our code will be released in \url{https://github.com/JD-AI-Research-NLP/P2DeNet}.}
% and \url{https://github.com/cuhksz-nlp}.}
\end{abstract}

\section{Introduction\label{intr}}
\noindent 
Natural language generation (NLG), aiming to automatically generate a sequence of tokens, are widely explored on tasks such as summarization, question answering and generation, dialog response generation, and machine translation.
Recently, generative pre-training models~\cite{radford2018improving,song2019mass,dong2019unified,lewis2019bart,raffel2019exploring,zhang2019pegasus,bi2020palm,xiao2020ernie,yan2020prophetnet}, which accumulate knowledge based on large-scale unsupervised conditional language modeling, have achieved remarkable improvements on downstream NLG tasks compared with conventional methods.
A typical generative pre-training model~\cite{song2019mass,lewis2019bart} follows the transformer~\cite{transformer} framework which contains an encoder and a decoder, where the decoder usually learns to generate a sequence in a left-to-right (L2R) order.
The L2R decoding strategy usually faces two issues during the modeling of target sequences: (1) limited to unidirectional context information, and (2) constrained on strong local dependencies. 

\begin{figure}[tb]
	\centering
	\includegraphics[width=2.9in]{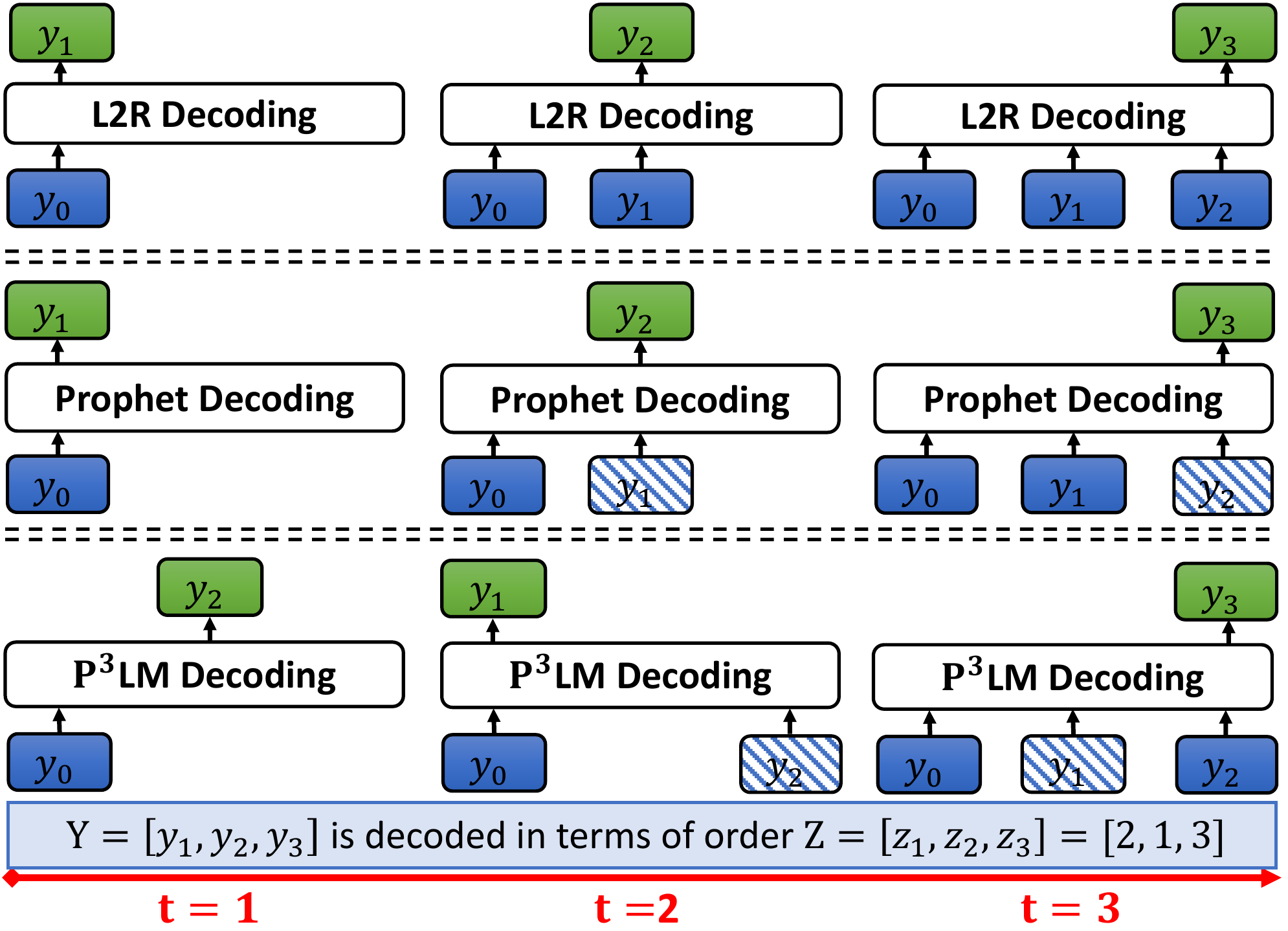}
% 	\caption{\label{task} An illustration of L2R decoding, prophet decoding, and P$^3$LM decoding. 
% 	\textbf{L2R decoding}: tokens \colorbox{cyan}{$y_{<t}$} are used to predict \colorbox{green}{$y_t$}. 
% 	\textbf{Prophet decoding}: tokens \colorbox{cyan}{$y_{<t}$} are used to predict both \colorbox{green}{$y_t$} and future token \colorbox{yellow}{$y_{t+1}$}. 
% 	\textbf{P$^3$LM decoding}: tokens \colorbox{cyan}{$y_{z_{<t}}$} are used to predict \colorbox{green}{$y_{z_{t}}$} and corresponding future token \colorbox{yellow}{$y_{z_{t+1}}$} with respect to sequence order $Z=[2,1,3]$.
	\caption{\label{task} An illustration of L2R, Prophet, and P$^3$LM decoding.
% 	of a sequence $Y=[y_1,y_2,y_3]$. 
	\textbf{L2R decoding}: {$y_t$} is predicted based on {$y_{\leq t-1}$}. 
	\textbf{Prophet decoding}: {$y_t$} is predicted based on {$y_{\leq t-1}$}, or {$y_{\leq t-2}$} with $y_{t-1}$ being masked.
	\textbf{P$^3$LM decoding}: $Y$ is autoregressively decoded in terms of order $Z$,
% 	$Z=[z_1,z_2,z_3]=[2,1,3]$, 
	where {$y_{z_{t}}$} is predicted based on $y_{z_{\leq t-1}}$, or $y_{z_{\leq t-2}}$ with $y_{z_{t-1}}$ being masked. $y_{0}\!=\!\langle s\rangle$ is the start of a sentence.
}
\vspace{-0.2cm}
\end{figure}

In order to enable a language model to learn bidirectional context information, auto-encoding ones, such as BERT~\cite{devlin2018bert} known as a masked language model (MLM), are pre-trained based on randomly masked token prediction.
In addition, autoregressive ones, such as XLNet\footnote{We clarify the differences between our P$^3$LM and XLNet in Appendix~\ref{sec:appendix_differ} in detail.}~\cite{yang2019xlnet} known as a permutation language model (PLM), are designed to reconstruct a partial sequence in permuted order.
However, directly applying these methods on language generation is not feasible, since they are designed for natural language understanding (NLU), which are usually handled by just one encoder or decoder~\cite{song2019mass}.
To prevent overfitting on strong local dependencies during decoding, ProphetNet~\cite{yan2020prophetnet} is proposed to predict $N$ future tokens.
However, the future token prediction strategy predicts at most $N$ (typically $N=2$) continuous tokens, which has limited ability on long dependency modeling.
Besides, due to the L2R decoding, the unidirectional target context modeling issue still exists.

To further enhance the ability of long dependency modeling,
% preliminarily addressed
as well as capturing bidirectional information of target sequences, 
% inspired by PLM, 
we propose P$^3$LM, a \textbf{p}robabilistically \textbf{p}ermuted \textbf{p}rophet \textbf{l}anguage \textbf{m}odel.
P$^3$LM learns to generate tokens in permuted order with an order-aware transformer decoder, as well as predicting the corresponding $N$ future tokens with a multi-stream attention mechanism.
Figure~\ref{task} illustrates the idea of the proposed P$^3$LM.
For instance, given a target sequence ${Y}=[y_1,y_2,y_3]$=\textit{sequence $\!\to\!$ order $\!\to\!$ matters} and a permuted order $Z=[2,1,3]$,
% where the $0$-th token is fixed as a special token $\langle s\rangle$ indicating the start of a sequence, 
P$^3$LM learns to generate sequence $Y$ in order $Z$, i.e., \textit{order $\!\to\!$ sequence $\to$ matters}.
Meanwhile, it also learns to predict future tokens in terms of $Z$, e.g., predicting \textit{sequence} as the future token of \textit{order} at time step $t=1$.
The above design makes P$^3$LM capable of capturing bidirectional information of target sequence, and strengths the modelling of long dependencies.
% For instance, at time step $t=2$, $y_1$ in green is predicted with $y_0$ and $y_2$ available.
% utilizes bidirectional context information and alleviates local token dependency in decoding procedure.
% to auto-regressively decode tokens in permuted order.
% We think URP decoding helps P$^3$LM to learn better relative positional information of tokens.
% Besides, equipped with the ability of predicting $n$ future tokens upon the specified order, P$^3$LM further strengths the modelling of bidirectional context and long token dependencies.

% In practice, P$^3$LM is implemented as the transformer encoder-decoder framework.
% , where an order-aware multi-stream decoding network is designed.
% Specifically, our model decodes tokens in permuted order with an order-aware masking strategy, and predicts future tokens with a multi-stream attention mechanism.
% Specifically, an order-aware transformer decoder is designed for permuted sequence generation, and a multi-stream attention mechanism is leveraged for $N$ future token prediction.
Extensive experiments are conducted on the GLGE~\cite{Liu2020GLGE} benchmark, a general language generation evaluation benchmark consisting of four datasets for summarization, two for question generation, one for conversational question answering, and one for dialog response generation, where our proposed P$^3$LM achieves 0.9 absolute and 2.5\% relative improvements on the overall score compared with the public available state-of-the-art model, i.e., ProphetNet.
% including CNNDM~\cite{see2017get}, Gigaword~\cite{rush2015neural}, and SQuAD-QG~\cite{du-etal-2017-learning}, where P$^3$LM achieves state-of-the-art results.
% Furthermore, P$^3$LM is proved to improve span-style machine reading comprehension (MRC) task on SQuAD-v1.1~\cite{rajpurkar2016squad} and SQuAD-v2.0~\cite{rajpurkar2018know}, in a generative way.
To conclude, the contributions are as follows:
% \begin{itemize}
(I) We propose P$^3$LM, a permutation over prophet decoding net, for generative pre-training, which utilizes bidirectional context information and enhances long token dependency modeling on target sequences;
(II) We conduct extensive experiments on downstream language generation tasks and show that P$^3$LM 
% achieves significant improvements and even 
obtains new state-of-the-art results on GLGE benchmark compared with published methods;
(III) Three P$^3$LM models, which cost about 100,000 dollars, are pre-trained based on large scale datasets and will be released for further research on generative pre-training and language generation for the NLP community.
% \end{itemize}
% Furthermore, a span-style MRC task is also verified to be improved by P$^3$LM in a generative way.
% Contribution or Advantages of P$^3$LM
%   \begin{itemize}
%       \item Combine Auto-regressive PLM and L2R within one unified framework. (1) P$^3$LM equals to Random-PLM + L2R; (2) Random-PLM equals to Uniform-PMLM.
%       \item Introducing PLM makes the model learn better position representation.
%       \item Context visible during decoding, GPT/BERT/RoBERTa/XL-Net is designed for NLU, like MASS and ProphetNet, P$^3$LM is designed for NLG
%       \item Unify prophet and permutation language models.
%   \end{itemize}
% which makes token prediction capable of observing bidirectional context
%我们提出用什么方法，面临什么挑战

%

% \begin{figure}[tb]
% 	\centering
% 	\includegraphics[width=3.0in]{figure/task.pdf}
% 	\caption{\label{task} An illustration.}
% \end{figure}

%##################################### Related Work ##########################################
% 是否可以给一个表格对比相关模型在模型架构（enc, dec, enc-dec）, 使用任务（NLU, NLG）, decoding order (L2R, R2L, Permuted Order), Long dependency (No, Yes)，Contextual info(unidirectional, bidirectional)。

% 可以把上述这个对比，跟图1合并起来。如果图1不能更易懂，可替换成表格。
\begin{table}[t]
	\centering
	\small
% 	\scriptsize
    \resizebox{0.48\textwidth}{!}{
	\begin{tabular}{llllll}
% 	\hline 
	\hline
	\specialrule{0em}{1pt}{1pt}
    \multirow{2}{*}{\bf Models}  & \multirow{2}{*}{\bf Structure} & \multirow{2}{*}{\bf Tasks}  &\multicolumn{3}{c}{\bf Features During Decoding} \\
    % \specialrule{0em}{1pt}{1pt}
    \cline{4-6}
    \specialrule{0em}{1pt}{1pt}
	& & & {\bf Order} & {\bf LongDep}&{\bf BiDir} \\
	\specialrule{0em}{1pt}{1pt}
	\hline
	\specialrule{0em}{1pt}{1pt}
	BERT & Enc & NLU & - & - & -\\
	RoBERTa & Enc & NLU & - & - & -\\
	XLNet & Enc & NLU & - & - & -\\
	ELECTRA & Enc & NLU & - & - & -\\
	ALBERT & Enc & NLU & - & - & -\\ 
% 	\specialrule{0em}{1pt}{1pt}
	\hline
	\specialrule{0em}{1pt}{1pt}
	GPT & Dec & NLU\&NLG & L2R & Shallow & No\\
	UniLM & Enc/Dec & NLU\&NLG & L2R\&R2L & Shallow & Shallow\\
	T5 & Enc-Dec & NLU\&NLG & L2R & Shallow & No\\
	BART & Enc-Dec & NLU\&NLG & L2R & Shallow & No\\ 
% 	\specialrule{0em}{1pt}{1pt}
	\hline
	\specialrule{0em}{1pt}{1pt}
	PEGASUS & Enc-Dec & NLG & L2R & Shallow & No \\
	PALM & Enc-Dec &  NLG & L2R & Shallow & No \\
	MASS & Enc-Dec & NLG & L2R & Shallow & No\\
	ProphetNet & Enc-Dec & NLG & L2R & Medium & No\\
	P$^3$LM & Enc-Dec & NLG & Permuted & Strong & Strong\\
% 	\specialrule{0em}{1pt}{1pt}
	\hline
% 	\hline
	\end{tabular}
	}
	\caption{\label{pretrain_works} Features about typical pre-trained models. \textbf{Enc}: encoder. \textbf{Dec}: decoder. \textbf{Order}$\in$\{L2R, R2L, Permuted\}: decoding order of target sequence. \textbf{LongDep}$\in$\{Shallow, Medium, Strong\}: long token dependencies in target sequence. \textbf{BiDir}$\in$\{No, Shallow, Strong\}: bidirectional information of target sequences. }
	\vspace{-0.1cm}
\end{table}
\section{Related Work}
\vspace{-0.1cm}
Typical pre-trained language models are shown in Table~\ref{pretrain_works}, which can be roughly classified into two categories: for natural language understanding (NLU) and for natural language generation (NLG).
Models~\cite{devlin2018bert,liu2019roberta,yang2019xlnet,lan2019albert,clark2020electra} that contain a single encoder, have been proved effective for dozens of downstream NLU tasks, e.g., XLNet \cite{yang2019xlnet} reconstruct a sentence fragment in permuted order.
% Different from the XLNet, P$^3$LM is a generative pre-training model with an encoder-decoder framework for NLG.
Another line of research is generative pre-training for NLG. 
Effective methods
% ~\cite{radford2018improving,song2019mass,dong2019unified,lewis2019bart,raffel2019exploring,zhang2019pegasus,yan2020prophetnet} 
have been designed to enhance NLG performance.
% which aim to enhance natural language generation.
These models usually pre-train the decoder as a left-to-right (L2R) autoregressive language model.
% GPT \cite{radford2018improving} pre-train a transformer-decoder in L2R order.
GPT-3~\cite{brown2020language} pre-train a transformer decoder with extremely large corpus and parameters, which is not finetuned on downstream tasks, while our model follows the pre-train then finetune framework.
% which can perform both NLU and NLG tasks. 
UniLM \cite{dong2019unified} pre-train a transformer encoder/decoder with both MLM task and sequence-to-sequence task, considering two unidirectional orders, i.e., L2R and R2L, while our model leverages permuted orders.
% However, both L2R and R2L are unidirectional and special cases of permuted order. 
Additional strong generative pre-trained models including MASS\cite{song2019mass}, BART\cite{lewis2019bart}, T5~\cite{raffel2019exploring}, and PEGASUS~\cite{zhang2019pegasus} utilize a transformer encoder-decoder framework to pre-train generative models, all of which are limited to train a L2R decoder, while our model learns to decode tokens in permuted order.
% T5 \cite{raffel2019exploring} explore huge data scale for generative pre-training.
% PEGASUS~\cite{zhang2019pegasus} explore  
% MASS \cite{song2019mass} pre-train a transformer encoder-decoder model with masked-sequence to sequence pairs where the decoder is only trained in an L2R order, while our model is enhanced with permuted order.
ProphetNet \cite{yan2020prophetnet} is the most similar approach to ours, which propose a future n-gram prediction mechanism for generative pre-training, while still limited to L2R decoding.
% In this paper, we propose a permutation over prophet (P$^2$) decoding network  the permutation and prophet ideas in one framework to enhance the modeling of future context and long distance token dependency for generation.

% \paragraph{Non-L2R Sequence Decoding}
% \cite{DBLP:journals/corr/VinyalsBK15}
% \cite{DBLP:journals/tacl/GuLC19}
% \cite{DBLP:conf/acl/LiaoJL20}
% \cite{DBLP:conf/nips/EmelianenkoVS19}
% \cite{DBLP:journals/corr/abs-1905-12790}
%################################# Approach #############################################
\section{Approach}
%################################ Model Overview #######################################
\subsection{Model Overview}
In this paper, probabilistically permuted prophet language modeling (P$^3$LM) is proposed for sequence generation.
The idea of P$^3$LM is learning to autoregressively generate a sequence in a probabilistically permuted order, meanwhile, multiple future tokens (in the perspective of that order) are jointly predicted at each decoding time step.
The above design of P$^3$LM makes it capable of capturing bidirectional information of a target sequence, as well as strengths the modelling of long dependencies in natural language.

% Preliminary: 
\subsubsection{Prophet Language Modeling}
To alleviate the problem of strong local dependencies during sequence generation, we introduce prophet decoding.
The original prophet modeling predicts $N$ words after current word. It is first utilized in Word2Vec~\cite{mikolov2013efficient}, where increasing range $N$ improves the word vector quality. ProphetNet~\cite{yan2020prophetnet} introduces it into sequence generation by predicting the future $N$ tokens.
Formally, given ${X}=[x_1,...,x_S]$ as a source sequence, and ${Y}=[y_1,...,y_{T}]$ as a target sequence.
The learning of a prophet language model (PLM) is to optimize the objective defined as follows:
\begin{equation}
\mathcal{L}_{plm}(Y|X)=\frac{1}{N}\sum_{n=1}^{N}\log p^{n}_{\theta}({Y}|{X})\nonumber\\
\end{equation}
where $\theta$ represents trainable parameters.
$p^{n}_{\theta}({Y}|{X})$ is the probability of generating ${Y}$ by skipping $n\in\{1,...,N\}$ tokens at each decoding time step $t$ defined as follows:
\begin{equation}
p^{n}_{\theta}({Y}|{X})=\prod_{t=1}^{T}p_{\theta}({y}_{t}|{y}_{{\leq t-n}},{X}) \nonumber
\end{equation}
In details, the prophet decoding can be viewed as a kind of masking strategy on previous generated sequence, namely, only $y_{\leq t-n}$ are feasible for predicting $y_{t}$.
In practice, to train models within reasonable computational complexity, $N$ is typically set as a small number, e.g., $N$ is $4$ for Word2Vec, and $2$ for ProphetNet.
This limits its ability of modeling long dependencies existing in natural language, such as long distance coreferences, clause dependencies, and discourse relations. Based on prophet decoding, we introduce P$^3$LM to address the problem in next section.

\subsubsection{P$^3$LM: Probabilistically Permuted Prophet Language Modeling}
Although prophet decoding is capable of alleviating the problem of strong local dependencies, its ability of long dependency modeling is still limited by small $N$ as described above, and it is constrained on unidirectional information due to L2R decoding.
The L2R order is a strong inductive bias, as it is natural for most human-beings to read and write sequences from left to right. 
Nevertheless, L2R is not the only option for generating sequences~\cite{DBLP:journals/tacl/GuLC19}.
For instance, people sometimes tend to think of central phrases first before building up a whole sentence.
Previous work has shown that order matters for sequence generation~\cite{DBLP:journals/corr/VinyalsBK15,DBLP:conf/nips/EmelianenkoVS19}.
Based on the above facts, a natural idea is to involve sequence order information into decoding.
To this end, we propose P$^3$LM to {strengthen prophet language model with probabilistically permuted sequence order}, which is capable of directly modeling long dependencies and bidirectional information of target sequences.
Formally, as previous study~\cite{DBLP:journals/corr/abs-1905-12790}, we condition the whole process on an input sequence $X$ to indicate that the proposed model is applicable to both conditional and unconditional sequence generation ($X=\varnothing$).
Specifically, the probability of generating ${Y}$ with prophet is defined as the expectation of its posterior probability $p^{n}_{\theta}({Y}|{X},{Z})$ over all possible orders as follows:
\begin{equation}
p^{n}_{\theta}({Y}|{X})=\mathbb{E}_{Z\sim p(Z)}p^{n}_{\theta}({Y}|{X},{Z}) \nonumber
\end{equation}
where order $Z=[z_1,...,z_T]\in P^{*}(T)$\footnote{$P^{*}(T)$ is the set of all permutations of $\{i\}_{i=1}^{i=T}$.}, which is a permutation of positions in $Y$, subjects to a prior distribution $p(Z)$. The decoding is further factorized according to order $Z$ as
\begin{equation}
p^{n}_{\theta}({Y}|{X},{Z})=\prod_{t=1}^{T}p_{\theta}({y}_{z_t}|{y}_{{z}_{\leq t-n}},{X}) \nonumber
\end{equation}
where ${y_{z_{t}}}$ represents the $t$-th generated token and $z_{t}$ is its absolute position in $Y$.
Training such a model needs to enumerate all the $T!$ permutations, which is impractical.
Instead, we maximize the lower bound $\mathcal{L}(Y|X)$ of the log likelihood $\mathcal{L}_{p^{3}lm}(Y|X)$ by sampling an order $\tilde{Z}$ according to the prior distribution $p(Z)$ as follows:
\begin{align}
    \mathcal{L}_{p^{3}lm}(Y|X)
    &\!=\!\frac{1}{N}\!\sum_{n=1}^{N}\log p^{n}_{\theta}({Y}|{X}) \nonumber \\
    &\!=\!\frac{1}{N}\!\sum_{n=1}^{N}\log \mathbb{E}_{Z\sim p(Z)}p^{n}_{\theta}({Y}|{X}\!,\!{Z}) \nonumber \\
    \geq\!\underbrace{\mathcal{L}(Y|X)}_{\text{lower bound}}\!&\!=\! \frac{1}{N}\!\sum_{n=1}^{N} [\log p^{n}_{\theta}({Y}|{X}\!,\!{\tilde{Z}})\!+\!\log p({\tilde{Z}})] \nonumber \\
    &\!=\!\frac{1}{N}\!\sum_{n=1}^{N}\sum_{t=1}^{T}\log p_{\theta}({y}_{\tilde{z}_{t}}|{y}_{\tilde{z}_{\leq t-n}}\!,\!{X})\!+\!{C} \nonumber 
    % &=\frac{1}{N}\sum_{n=0}^{N\!-\!1}\sum_{t=1}^{T}\log \underbrace{p_{\theta}({y}_{z^{o}_{t}}|{y}_{{z}^{o}_{<t-n}},{X})}_{\text{P$^3$LM encoder-decoder}}+\underbrace{C}_{\text{const}} \nonumber 
    \nonumber
\end{align}
Theoretically, different distribution $p(Z)$ will result in different P$^3$LMs.
Exploring the best distribution $p^{*}(Z)$ could be an interesting problem for future research. 
In this paper, we preliminarily explore an $\alpha$-P$^3$LM which combines L2R and URP distributions which are defined as follows:
\begin{itemize}
    \item {\bf L2R} order ${Z}^{\mathtt{L2R}}=[1,...,T]$ is the {l}eft-to-{r}ight position sequence of words in ${Y}$. Most previous methods train a model to generate target sequences in L2R order. The corresponding $p(Z)$ of these methods, which is a pulse distribution, is defined as follows:
    \begin{equation}
        p^{\mathtt{L2R}}(Z)=
        \begin{cases}
        1,& Z=Z^{\mathtt{L2R}} \\
        0,& Z\neq Z^{\mathtt{L2R}}
        \end{cases} \nonumber
    \end{equation}

    \item {\bf URP} order means an uniformly random permutation of the word positions in ${Y}$. The corresponding $p(Z)$, which is an uniform distribution over the $T!$ permutations $P^{*}(T)$, is defined as follows:
    \begin{equation}
        p^{\mathtt{URP}}(Z)=\frac{1}{T!},~~~ Z\in P^{*}(T) \nonumber
    \end{equation}
\end{itemize}
% In contrast, we consider both L2R and URP distributions for sequence generation.
% The motivation is that 
We believe that the diverse URP orders (not only L2R) can help strengthen the modeling of bidirectional information and long dependencies of target sequences.
Finally, the order distribution of $\alpha$-P$^3$LM is straightforwardly defined as
% follows:
% Therefore, we finally define an $\alpha$-P$3$LM with the order distribution straightforwardly defined as the trade-off of L2R and URP distributions as follows:
$$p^{\alpha}(Z)=\alpha p^{\mathtt{L2R}}(Z)+(1-\alpha)p^{\mathtt{URP}}(Z)$$
In this paper, we empirically set $\alpha=0.5$ according to experiments. 
Besides, unlike previous works that focus on automatically determining a best generation order during inference~\cite{DBLP:journals/tacl/GuLC19,DBLP:conf/nips/EmelianenkoVS19,DBLP:journals/corr/abs-1905-12790} which requires nontrivial design, we focus on modeling orders during training and keep the L2R inference to reduce the complexity of the model. 

%===============================================================================================================
\begin{figure*}[t]
	\centering
	\includegraphics[width=6.2in]{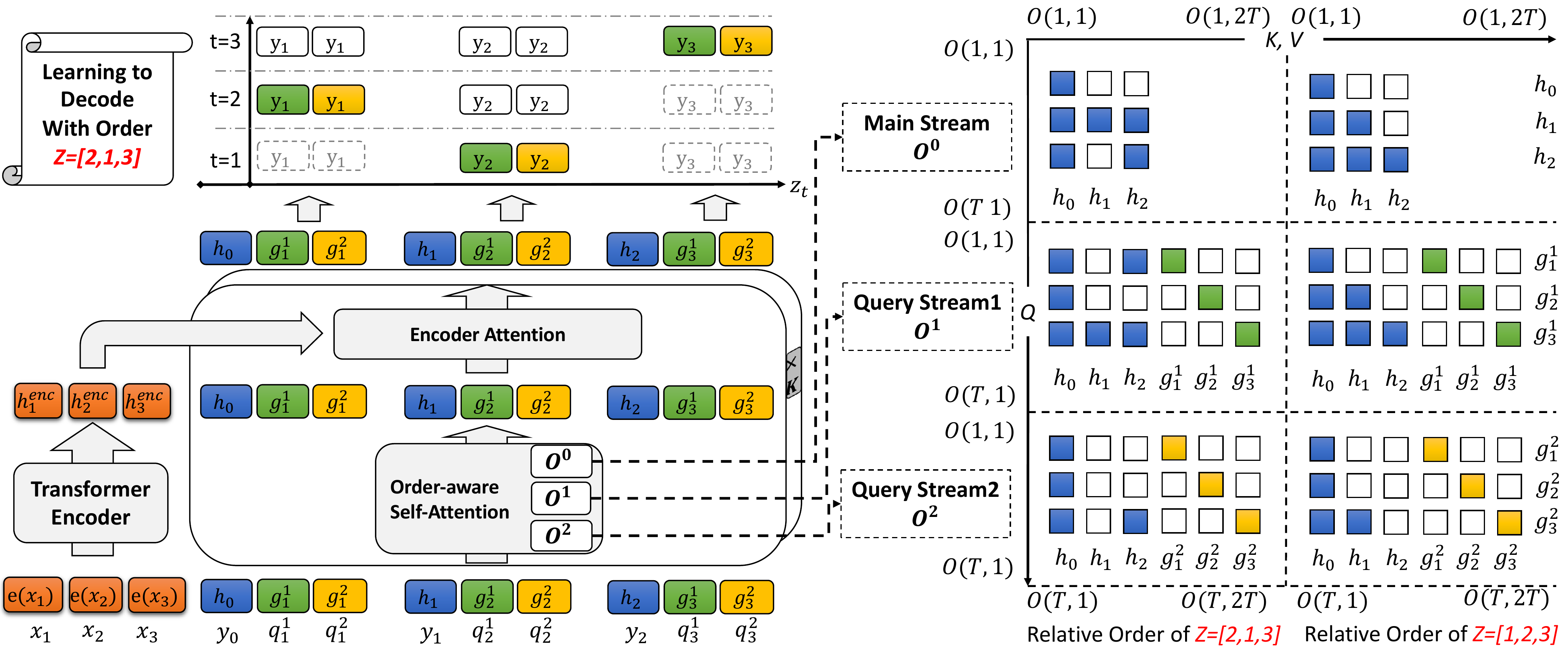}
	\caption{\label{model} Neural architecture of P$^3$LM (Left), and sampled relative orders (Right) as examples.
	A case is shown that sequence $y_2 \to y_1 \to y_3$ is decoded in order $Z=[2,1,3]$ where $y_0=\langle s \rangle$ is fixed as the start of sentence. Encoder is in {\color{red}red}, main stream is in {\color{blue}blue}, query stream 1 is in {\color{green}green}, and query stream 2 is in
 	\textcolor[RGB]{255,128,0}{yellow}. Colored items in the masking tensors mean that the corresponding input is available for computing the corresponding output.
	}
\end{figure*}

\subsection{Neural Architecture of P$^3$LM}
The backbone of the proposed P$^3$LM is a transformer encoder-decoder~\cite{transformer} illustrated in Figure~\ref{model}.
The encoder transfers ${X}$ into hidden states
%%%%%%%%%%%%%%%%%%%%%%%%%%
$\mathbf{h}^{e}\!=\!\mathtt{enc}({X})$
%%%%%%%%%%%%%%%%%%%%%%%%%%
where $\mathtt{enc}(\cdot)$ is a standard transformer encoder.
According to the objective $\mathcal{L}(Y|X)$ defined above, during training, the P$^3$LM decoder \textit{simultaneously} calculates $T\!\times\!N$ probabilities as follows:
$$\{p_{\theta}({y}_{{z}_{t}}|{y}_{{z}_{\leq t-n}},{X})\}_{t=1,n=1}^{t=T,n=N}\!=\!\mathtt{dec}(Y\!,\!Z\!,\!N\!,\mathbf{h}^{e})$$
Compared with the vanilla transformer decoder, our decoder $\mathtt{dec}(\cdot)$ has two characteristics: (1) it takes an order $Z$ as additional input to guide the autoregressive generation, and (2) it can simultaneously skip $[1,...,N]$ (Note that $N\!=\!1$ means the next token prediction) \textit{previous} tokens for prediction at each time step.
To achieve the above two capabilities, 
% inspired by XLNet~\cite{yang2019xlnet} and ProphetNet~\cite{yan2020prophetnet}, 
we implement an \textit{order-aware multi-stream} P$^3$LM decoder with its workflow shown in Algorithm~\ref{algo}.
The major effort of such a decoder is to model the absolute order $Z$ as \textit{\textbf{relative orders}} in multi-stream attention, aiming to control what information to use or not for decoding.
% Note that, as we described above, left-to-right decoding is used during inference, i.e., $Z\!=\!Z^{\mathtt{L2R}}$ and $N\!=\!1$.

\begin{algorithm}[t]
\caption{P$^3$LM Decoder $\mathtt{dec}(\cdot)$}
\label{algo}
\SetKwData{Index}{Index}
% \LinesNumbered 
\KwIn{Target sequence $Y$, order $Z$, place holders $\{q^{n}_{z_{t}}\}_{t=1,n=1}^{t=T,n=N}$, \# of query streams $N$, encoder hidden states $\mathbf{h}^{e}$}
% \KwIn{Target sequence $Y$, order $Z$, place holders $\{q^{n}_{z_{t}}\}_{t=1,n=1}^{t=T,n=N}$, \# of query streams $N$, encoder hidden states $\mathbf{h}^{e}$}
\KwOut{
Prediction probabilities 
$\{p_{\theta}({y}_{{z}_{t}}|{y}_{{z}_{\leq t-n}}\!,\!{X})\}_{t=1,n=1}^{t=T,n=N}$}
% \textbf{def} order2mask(Z):\\
Create $0$-initialized tensors $O^{0}\in \mathbb{R}^{T\times T}$ and $O^{1},...,O^{N}\in \mathbb{R}^{T\times 2T}$ as relative orders\\
Set $O^{*}(*,1)=1$\\
% \For {$j$ in $[1,...,T]$, $i$ in $[1,...,T]$}
\For {$i\leftarrow 1$ \KwTo $T$, $j\leftarrow 1$ \KwTo $T$}
{
    let $z_{t1}=i$ and $z_{t2}=j$\\
    % $\underbrace{M^{0}(j+1,i+1)}_{i\neq T,j\neq T}=1$ if $t1\leq t2$\\
    ${O^{0}{(j\!+\!1,i\!+\!1)}}\!=\!1$ \textbf{if} $t1\!\leq\!t2,i,j\!\neq\!T$\\
    % \begin{cases}
    %     1,& t1\leq t2 \\
    %     0,& \text{others}
    % \end{cases} \nonumber\\

    \For{$n\leftarrow 1$ \KwTo $N$}
    {
        % $\underbrace{M^{n}(j,i+1)}_{i\neq T}=1$ if $t1\leq t2-n$\\
        ${O^{n}{(j,i\!+\!1)}}\!=\!1$ \textbf{if} $t1\!\leq\! t2\!-\!n,i\!\neq\! T$\\
        % \begin{cases}
        %     1,& t1\leq t2-n \\
        %     0,& \text{others}
        % \end{cases} \nonumber\\
        ${O^{n}{(j,i\!+\!T)}}\!=\!1$ \textbf{if} $t1=t2$ \\
        % \begin{cases}
        %     1,& t1=t2 \\
        %     0,& \text{others}
        % \end{cases} \nonumber\\
    }
}
embed $[\mathtt{\langle s\rangle},Y]$ as $\mathbf{h}$ and $q^{n}$ as $\mathbf{g}^{n}$, where $q^{n}\!=\!\{q^{n}_{z_{t}}\}_{t=1}^{t=T}$ and $\mathbf{g}^{n}\!=\!\{\mathbf{g}^{n}_{z_{t}}\}_{t=1}^{t=T}$\\
\For{$k\leftarrow 1$ \KwTo $K$}
{
    $\mathbf{h} \leftarrow \mathtt{OSA}_{\phi^{0}}(\mathbf{h},\mathbf{h},\mathbf{h},O^{0})$\\
    $\mathbf{h} \leftarrow \mathbf{h}$ encoder attention on $\mathbf{h}^{e}$\\
    \For{$n\leftarrow 1$ \KwTo $N$}
    {
        % $\mathbf{hg}^{n}=[\mathbf{h};\mathbf{g}^{n}]$\\
        $\mathbf{g}^{n}\! \leftarrow\! \mathtt{OSA}_{\phi^{n}}(\mathbf{g}^{n},\![\mathbf{h};\!\mathbf{g}^{n}],[\mathbf{h};\!\mathbf{g}^{n}],\!O^{n})$\\
        $\mathbf{g}^{n} \leftarrow \mathbf{g}^{n}$ encoder attention on $\mathbf{h}^{e}$\\
    }
}
return $p^{n}_{\theta}({y}_{{z}_{t}}|{y}_{{z}_{\leq t-n}}\!,\!X)\!=\!\mathtt{softmax}(\mathbf{g}^{n}_{{z}_{t}}W)$
\end{algorithm}
% \subsubsection{Order-Aware Multi-Stream Decoder}
\paragraph{Multi-Stream.} The original multi-stream attention has been successfully utilized in XLNet~\cite{yang2019xlnet}. 
Different from XLNet which leverages a 2-stream 
% (one main stream for representation, and a single query stream for prediction) 
attention in \textit{encoder for NLU} tasks, we adopt an $(N\!+\!1)$-stream attention where $N\!\geq\!1$ in \textit{decoder for NLG} tasks like ProphetNet~\cite{yan2020prophetnet}.
Unlike ProphetNet, attention in P$^3$LM decoder is {order sensitive}.
Specifically, as shown in Figure~\ref{model} and lines 9-16 in Algorithm~\ref{algo}, at each time step $t$, P$^3$LM decoder leverages a main stream (in blue) as the vanilla transformer decoder to represent $y_{z_{<t}}$ as hidden states $\mathbf{h}_{{z}_{<t}}$.
In addition, it constructs $N$ query streams (in green and yellow) to represent $N$ place holders $q_{z_{t}}\!=\![q_{z_{t}}^{1},...,q_{z_{t}}^{N}]$ as hidden states $\mathbf{g}_{{z}_{t}}\!=\![ \mathbf{g}^{1}_{{z}_{t}},...,\mathbf{g}^{N}_{{z}_{t}}]$.
Each query stream is used to predict $y_{z_t}$ by skipping $n \in [1,...,N]$ tokens, respectively.
The above multi-stream transformation is implemented with $K$ layers (line 10), each of which contains two sub-layers, i.e., order-aware self-attention $\mathtt{OSA}(\phi)$ (line 11 and 14) introduced in next section and encoder-attention (line 12 and 15).
% Therefore, different streams are updated according to Algorithm~\ref{algo}.
Finally, the distribution of predicting ${y}_{{z}_{t}}$ at the $n$-th stream is defined as
%%%%%%%%%%%%%%%%%%%%%%%%%%
$p^{n}_{\theta}({y}_{{z}_{t}}|{y}_{{z}_{<t-n}},{x})=\mathtt{softmax}(\mathbf{g}^{n}_{{z}_{t}}W)$ (line 16)
%%%%%%%%%%%%%%%%%%%%%%%%%%
where $W \in \mathbb{R}^{D \times V}$ are trainable parameters, $D$ indicates the hidden size, and $V$ represents the vocabulary size.
% , as the vanilla transformer decoder. 
% The difference is that the self-attention in P$^3$LM decoder is \textit{order-aware}.
% defined as follows.
% The decoder then further transfers the above hidden states with $K$ order-aware multi-stream attention layers, where each layer is consist of an \textbf{o}rder-aware \textbf{m}ulti-stream \textbf{s}elf-\textbf{a}ttention (OMSA) layer and an \textbf{o}rder-aware \textbf{m}ulti-stream \textbf{c}ross-\textbf{a}ttention (OMCA) layer.

% \textbf{OM-SA} layer encodes hidden states from the $k-1$ layer and calculates the main stream and query stream hidden states at the $k$-th layer as follows:
% \textbf{OMSA} layer transforms hidden states from the $k-1$ layer and calculates the main stream and query stream hidden states at the $k$-th layer as follows:

\paragraph{Order-aware Self-Attention (OSA).} 
To involve the order information, an intuitive solution is to directly reorder $Y$ into a new sequence $Y'$ according to $Z$, and then learns to decode $Y'$ with an L2R decoder.
However, it will mismatch word and position embeddings, which leads to the loss of the words' original positional information.
Instead, we introduce an order-aware self-attention $\mathtt{OSA}_{\theta}(\cdot)$ which leverages relative orders and keeps the positions of the words inputted into the decoder unchanged.
Specifically, the absolute order $Z$ is converted into relative order $O^{0}\in \mathbb{R}^{T\times T}$ for main stream, and a set of relative orders $O^{1},...,O^{N}\in \mathbb{R}^{T\times 2T}$ for query streams (lines 1-8).
$O(j,i)$ indicates the item in the $j$-th row and $i$-th column of a matrix $O$.
In short, these relative orders act as attention masks, controlling that words with their order in front are available for those behind.
Finally, $\mathtt{OSA}_{\phi}(\cdot)$ taking in packed hidden states $Q,K,V$ and some relative order $O$ is defined as follows:
%%%%%%%%%%%%%%%%%%%%%%%%%%
% \begin{align}
%     % \mathbf{h}\leftarrow
%     &\mathtt{OMA}_{\theta}(\underbrace{Q}_{{Q}\in \mathbb{R}^{T\times D}},\underbrace{K,V}_{{K,V}\in \mathbb{R}^{2T\times D}},\underbrace{M}_{{M}\in \mathbb{R}^{T\times 2T}}) \nonumber \\
%     =&\mathtt{softmax}(\frac{(QW^{Q})\cdot (KW^{K})^{\top}\odot {M}}{\sqrt{D}})\cdot (VW^{V}) \nonumber
%     % =&\mathtt{softmax}(\frac{QK^{T}\!\odot\!M}{\sqrt{D}})V \nonumber
% \end{align}
\begin{align}
    % \mathbf{h}\leftarrow
    &\mathtt{OSA}_{\phi}({Q},{K,V},{O}) \nonumber \\
    =&\mathtt{softmax}(\frac{(QW^{Q})\cdot (KW^{K})^{\top}\odot {O}}{\sqrt{D}})\cdot (VW^{V}) \nonumber
    % =&\mathtt{softmax}(\frac{QK^{T}\!\odot\!M}{\sqrt{D}})V \nonumber
\end{align}
where $W^{Q},W^K,W^V\!\in\! \phi$ are trainable parameters.

%############################## Experiment ########################################
\section{Experiment}
Extensive experiments are conducted.
In Section~\ref{p2decpre-train}, pre-training details of P$^3$LM are introduced.
% Besides, we show that more pre-training iterations improve downstream tasks.
In Section~\ref{Generation}, we show that P$^3$LM achieves state-of-the-art (SOTA) results on GLGE benchmark compared with published methods.
% including text summarization, question generation, conversational question answering, and persona dialog response generation, compared with published methods.
In Section~\ref{Summarization}, we conduct experiments on text summarization dataset CNN/DM, where
% , and show the effectiveness of our P$^3$LM.
% Furthermore, 
ablation study verifies the effectiveness of P$^3$LM which involves sequence order information compared with conventional {l}eft-to-{r}ight (L2R) generation paradigm.
% , no matter with or without pre-training.

%############################## P$^3$LM Pre-training ##################################
\subsection{P$^3$LM Pre-training\label{p2decpre-train}}
%############################## Model Architecture #################################
\subsubsection{Model Architecture}
P$^3$LM follows the transformer encoder-decoder framework.
Two model architectures, i.e., P$^3$LM$_{\mathtt{base}}$ and P$^3$LM$_{\mathtt{large}}$ are used for pre-training.
The base architecture contains about 125M parameters including a 6-layer encoder and a 6-layer decoder with 768 embedding/hidden size and 3,072 feed-forward filter size.
The architecture of the large model contains about 391M parameters including a 12-layer encoder and 12-layer decoder with 1,024 embedding/hidden size and 4,096 feed-forward filter size.
% The batch size and training steps are set to 1,024 and 1,500K, respectively.

%############################# Pre-Training Dataset, Infrastructure, and details ##############################
\subsubsection{Corpus and Infrastructure} %Pre-Training 
Following BERT and ProphetNet~\cite{yan2020prophetnet}, the English Wikipedia and BookCorpus are used to pre-train P$^3$LM.
In this paper, to keep up with previous work, we first collect and process the above datasets, and finally obtain about 16GB data for pre-training.
We pre-train a P$^3$LM$_{\mathtt{base(16G)}}$ and a P$^3$LM$_{\mathtt{large(16G)}}$ model on the 16GB dataset with 64 $\times$ 32GB NVIDIA V100 GPUs from scratch without loading any other pre-trained models.
Following ProphetNet on large scale pre-training, we also collect a 160GB large scale dataset which is the combination of five sources including wikipedia, books, stories, news, and web text.
Based on the 160G data, we also pre-train a large scale model P$^3$LM$_{\mathtt{large(160G)}}$ initialized by P$^3$LM$_{\mathtt{large(16G)}}$ with 16 $\times$ 40GB NVIDIA A100 GPUS.
The batch size of all the three pre-trained models are set as 1,024.
The P$^3$LM$_{\mathtt{base(16G)}}$, P$^3$LM$_{\mathtt{large(16G)}}$, and P$^3$LM$_{\mathtt{large(160G)}}$ are trained with 750k (95 epochs), 1,500k (192 epochs), and 2000k (22 epochs) iterations and cost about 1.7 days, 28.0 days, 48.6 days, respectively.
We use Adam optimizer~\cite{kingma2014adam} with a learning rate of 1e-4 for pre-training.
Our implementation is based on FAIRSEQ\footnote{\url{https://fairseq.readthedocs.io/en/latest/}}.
To make a fair comparison, we set the maximum future $N$-token to be $2$ as ProphetNet in experiments.

%#################################### Pre-Training Task #####################################

\subsubsection{Pre-Training Task}
P$^3$LM is essentially a sequence-to-sequence model which takes a sequence as input and outputs a target sequence.
During pre-training, the input length is set to 512 tokens.
We randomly pick a starting position $u$ in every 64 tokens, and then mask a continuous span from $u$.
The masked length is set to 15\% of the total number of input tokens, i.e., 9 continuous tokens in every 64 tokens.
Following ProphetNet and MASS~\cite{song2019mass}, among the masked tokens, 80\% of them are replaced by $[\mathbb{M}]$, 10\% replaced by random tokens, and 10\% unchanged.
Considering the computational cost, we follow MASS to only predict the masked fragment.
Different from ProphetNet and MASS, P$^3$LM predicts the target sequence in both an L2R order and a URP order.
Specifically, a URP sequence generation task is to generate a target sequence word by word in a given URP sequence order.
Traditional L2R sequence generation task trains a generative model which only needs to learn a fixed \textit{one-word-right} relative positional information.
In contrast, URP sequence generation requires a model to learn more complex \textit{arbitrarily} relative positional information between words in a target sequence.

\subsection{Finetune on General Generation Tasks\label{Generation}}
In this section, we show the finetune results of P$^3$LM compared with strong baselines and state-of-the-art pre-trained models on GLGE\footnote{\url{https://microsoft.github.io/glge/}}~\cite{Liu2020GLGE}, which is a general language generation evaluation benchmark containing 8 datasets on 4 tasks. 
% Details are in Appendix~\ref{sec:appendix_glge}.

\begin{table*}[t]
	\centering
% 	\small
% 	\scriptsize
    \resizebox{\textwidth}{!}{
	\begin{tabular}{lrcccccccc}
	\hline \hline
% 	\toprule
	\multirow{2}{*}{\bf Models} &\multirow{2}{*}{\bf Score} & \multicolumn{4}{c}{\bf Text Summarization} & \multicolumn{2}{c}{\bf Question Generation}& \multicolumn{1}{c}{\bf QA}& \multicolumn{1}{c}{\bf Dialog} \\
% 	\cr	\cmidrule(lr){3-6} \cmidrule(lr){7-8} \cmidrule(lr){9} \cmidrule(lr){10} 
	&&{\bf CNN/DM}&{\bf Gigaword}&{\bf XSUM}&{\bf MSNews}&{\bf SQuAD1.1}&{\bf MSQG}&{\bf CoQA}&{\bf PersonaChat} \\
	\hline
	Metrics && \multicolumn{4}{c}{ R-1/R-2/R-L} & \multicolumn{2}{c}{R-L/B-4/MTR}& \multicolumn{1}{c}{F1}& \multicolumn{1}{c}{B-1/B-2/D-1/D-2} \\
	\hline
	\multicolumn{10}{c}{\bf Test} \\
	\hline
	LSTM & \colorbox{lightgray}{20.0} & 37.3/15.7/34.4 & 34.2/16.0/31.8 & 25.1/6.9/19.9 & 30.0/14.6/27.7 & 27.2/3.8/8.9 & 25.3/3.5/14.1 & 15.1 & 42.2/35.9/0.2/0.7\\
	
	Transformer & \colorbox{lightgray}{21.9}& 39.5/16.7/36.7 & 37.1/18.4/34.5 & 30.5/10.4/24.2 & 33.0/15.4/30.0 & 30.7/4.8/10.9 & 29.3/5.1/16.6 & 15.7 & 38.3/33.6/0.2/0.7\\
	
	MASS$_{\mathtt{base}}$ & \colorbox{lightgray}{33.6}& 42.1/19.5/39.0 & 38.7/19.7/35.9 & 39.7/17.2/31.9 & 39.4/21.0/36.1 & 49.4/20.1/24.4 & 38.9/10.2/23.3 & 65.4 & 41.0/35.7/1.4/6.9\\
	
	ProphetNet$_{\mathtt{base}}$ & \colorbox{lightgray}{33.8}& 42.5/19.7/39.5 & 38.9/19.9/36.0 & 39.8/17.1/32.0 & 40.6/21.6/37.0 & 48.0/19.5/23.9 & 37.1/9.3/22.7 & 65.3 & 46.0/38.4/1.3/7.3\\
	
	MASS$_{\mathtt{middle}}$ & \colorbox{lightgray}{34.3}& 42.9/19.8/39.8 & 38.9/20.2/36.2 & 39.1/16.5/31.4 & 40.4/21.5/36.8 & 49.9/21.3/25.2 & 38.9/9.5/23.5 & 67.6 & 46.0/38.2/1.2/6.2\\
	
	BART$_{\mathtt{large}}$ & \colorbox{lightgray}{35.8}& 44.1/{\bf 21.2}/40.9 & 38.1/18.4/34.9 & {45.1}/{22.2}/{37.2} & 43.8/24.0/39.2 & 50.3/22.0/26.4 & 38.8/9.2/{\bf 24.3} & 68.6 & {\bf 49.9}/{\bf 40.0}/1.3/8.0\\
	
	ProphetNet$_{\mathtt{large}}$ & \colorbox{lightgray}{36.5} & 44.2/21.1/41.3 & 39.5/{\bf 20.4}/{36.6} & 44.4/21.3/36.4 & 44.1/24.4/40.2 & 51.5/22.5/26.0 & 38.3/9.6/23.3 & 73.0 & 46.7/39.0/1.3/7.5\\
	
	P$^3$LM$_{\mathtt{large(160G)}}$ & \colorbox{lightgray}{\bf 37.4} & {\bf 44.3}/21.0/{\bf 41.4} &	{\bf 39.6}/20.2/{\bf 36.8} &	{\bf 45.3}/{\bf 22.3}/{\bf 37.3} &	{\bf 44.6}/{\bf 25.0}/{\bf 40.8} &	{\bf 51.6}/{\bf 23.0}/{\bf 26.6} &	{\bf 39.5}/{\bf 11.0}/23.6 &	{\bf 75.3} &	48.8/39.4/{\bf 1.7}/{\bf 13.7}\\
	
% 	P$^3$LM$_{\mathtt{large(160G)}}$ & \colorbox{lightgray}{\bf 37.4} & {\bf 44.3}/21.0/{\bf 41.4} &	{\bf 39.6}/20.2/{\bf 36.8} &	{\bf 45.3}/{\bf 22.3}/{\bf 37.3} &	{\bf 44.6}/{\bf 25.0}/{\bf 40.8} &	{\bf 51.6}/{\bf 23.0}/{\bf 26.6} &	{\bf 39.5}/{\bf 11.0}/23.6 &	{\bf 75.3} &	47.9/39.1/{\bf 1.8}/{\bf 13.6}\\

	\hline
	\multicolumn{10}{c}{\bf Valid} \\
	\hline
	
	P$^3$LM$_{\mathtt{large(160G)}}$ & \colorbox{lightgray}{38.7} & 44.8/21.5/42.0 &	48.8/26.8/45.6 &	45.4/22.5/37.6 &	44.2/24.5/40.4 &	52.5/24.3/27.1 &	39.9/12.9/24.4 &	75.9  &	49.0/39.4/1.7/13.5 \\
% 	P$^3$LM$_{\mathtt{large(160G)}}$ & \colorbox{lightgray}{38.7} & 44.8/21.5/42.0 &	48.8/26.8/45.6 &	45.4/22.5/37.6 &	44.2/24.5/40.4 &	52.5/24.3/27.1 &	39.9/12.9/24.4 &	75.9  &	48.2/39.0/1.7/13.3 \\

	\hline \hline
	\end{tabular}
	}
	\caption{\label{cnndm_160g} Experiment results on the GLGE. Overall scores $s=\frac{1}{8}\sum_{d=1}^{8}\!\frac{1}{|S_{d}|}\!\sum_{m\in S_{d}}\!m$ where $S_d$ is the metrics for the $d$-th dataset are highlighted in \colorbox{gray}{color}. R-1: Rouge-1. R-2: Rouge-2. R-L: Rouge-L. B-4: BLUE-4. MTR: METEOR. D-1: Distinct-1. D-2:Distinct-2. It is worth noting that D-1 and D-2 are multiplied by 100. Highest scores are in bold. The gains of P$^3$LM$_{\mathtt{large(160G)}}$ over ProphetNet$_{\mathtt{large(160G)}}$ are statistically significant at $p = 0.05$.}
\end{table*}

\begin{table}[h]
	\centering
	\small
% 	\scriptsize
    \resizebox{0.47\textwidth}{!}{
	\begin{tabular}{lrrrrr}
	\hline \hline
	{\bf Corpus} & {\bf $|$Train$|$} & {\bf $|$Dev$|$} & {\bf $|$Test$|$} & {\bf $|$Src.$|$}&{\bf $|$Tgt.$|$} \\
	\hline
	CNN/DM & 287,113 & 13,368 & 11,490 & 822.3 & 57.9\\
	Gigaword & 3,803,957 & 189,651 & 1,951 & 33.7 & 8.7\\
	XSUM & 204,017 & 11,327 & 11,333 & 358.5 & 21.1\\
	MSNews & 136,082 & 7,496 & 7,562 & 310.7 & 9.7\\
	SQuAD 1.1 & 75,722 & 10,570 & 11,877 & 149.4 & 11.5\\
	MSQG & 198,058 & 11,008 & 11,022 & 45.9 & 5.9\\
	CoQA & 108,647 & 3,935 & 4,048 & 354.4 & 2.6\\
	PersonaChat & 122,499 & 14,602 & 14,056 & 120.8 & 11.8\\
	\hline \hline
	\end{tabular}
	}
	\caption{\label{GLGE_statistic} Statistics of GLGE tasks. $|$Train$|$: the number of examples in training set. $|$Src.$|$: the average number of words in source inputs.}
\end{table}

\subsubsection{GLGE Benchmark}
\label{sec:appendix_glge}
Table~\ref{GLGE_statistic} shows the statistics of GLGE benchmark.
GLGE is a general language generation evaluation benchmark consisting of four datasets for summarization including CNN/DM \cite{hermann2015cnndm,see2017get}, Gigaword \cite{rush2015neural,graff2003gigaword}, XSum \cite{narayan2018don}, and MSNews, two for question generation including SQuAD 1.1 \cite{rajpurkar2016squad}, and MSQG, one for conversational question answering including CoQA \cite{reddy2019coqa}, and one for dialog response generation including PersonaChat \cite{zhang2018personalizing}.
Statistics of GLGE are show in Table~\ref{GLGE_statistic}.
Evaluation metrics, including Rouge-1, ROUGE-2, and ROUGE-L~\cite{lin2004rouge} for summarization,
ROUGE-L, BLEU-4~\cite{papineni2002bleu}, and METEOR~\cite{banerjee2005meteor} for question generation, F1 for conversational question answering, and BLEU-1, BLEU-2, Distinct-1, and Distince-2~\cite{li2015diversity} for dialog response generation, are used.
GLGE calculates an overall score $s=\frac{1}{8}\sum_{d=1}^{8} \frac{1}{|S_{d}|} \sum_{m\in S_{d}}m$ where $S_d$ indicates the evaluation metrics for the $d$-th dataset.

% \subsubsection{Baselines}
% We choose the following well performed pre-trained generative models as our baselines including {LSTM}~\cite{bahdanau2014neural}, {Transformer}~\cite{transformer}, {MASS}~\cite{song2019mass}, {BART$_{\mathtt{large}}$}~\cite{lewis2019bart}, and {ProphetNet}~\cite{yan2020prophetnet}. The details of these models are introduced in Appendix~\ref{sec:appendix_baselines}.
% % \begin{itemize}
% % \item 

\subsubsection{Baselines}
We choose the following well performed pre-trained generative models as our baselines.
% \label{sec:appendix_baselines}
\textbf{LSTM}~\cite{bahdanau2014neural} is implemented with the word embedding dimension, the hidden size, the number of the encoder layer, and the number of the decoder layer as 512, 512, 1, and 1, respectively. 
% Adam with an initial learning rate of between 1e-4 and 3e-4 is used, and 
LSTM is trained for a maximum of 100 epochs with learning rate of between 1e-4 and 3e-4.
%
% \item 
\textbf{Transformer}~\cite{transformer} contains a 6-layer encoder and a 6-layer decoder with 1024 embedding and hidden size, and 4096 feed-forward filter size.
% Adam with the initial learning rate of between 1e-4 and 3e-4 is used, and 
Transformer is trained for a maximum of 20 epochs with learning rate of between 1e-4 and 3e-4.
% 
% \item 
\textbf{MASS}~\cite{song2019mass} includes MASS$_{\mathtt{base}}$ and MASS$_{\mathtt{middle}}$ containing a 6-layer encoder and a 6-layer decoder with 768/1024 embedding and hidden size and 3072/4096 feed-forward filter size. MASS are pre-trained on the 16GB English Wikipedia and BookCorpus dataset and finetuned for a maximum of 25 epochs.
% 
% \item 
\textbf{BART$_{\mathtt{large}}$}~\cite{lewis2019bart} contains a 12-layer encoder and 12-layer decoder with 1024 embedding and hidden size, and 4096 feed-forward filter size. 
BART is pre-trained based on the 160GB data of news, books, stories, and web text and finetuned for a maximum of 20,000 iterations.
% 
% \item 
\textbf{ProphetNet}~\cite{yan2020prophetnet} includes ProphetNet$_{\mathtt{base}}$ and ProphetNet$_{\mathtt{large}}$ containing the same architecture as the corresponding P$^3$LM models, where the base model is pre-trained on the 16GB English Wikipedia and BookCorpus, and the large one on the 160GB corpora. ProphetNet is finetuned for a maximum of 10 epochs.
% We do not compare with CTRLgen since it is not published and we can not implement it.
% \end{itemize}

\subsubsection{Implementation Details}
{P$^3$LM$_{\mathtt{large(160G)}}$} is pre-trained on the same 160GB data as ProphetNet$_{\mathtt{large}}$ as described in Section~\ref{p2decpre-train}, and then finetuned on the eight datasets in GLGE, respectively.
The best performing model on each development set is chosen to inference on the corresponding test set.
Due to space limitation, implementation details about the eight models are show in Table~\ref{P2DeNET_imp} in Appendix~\ref{sec:appendix_param}.

\subsubsection{Main Results}
Table~\ref{cnndm_160g} shows the results of P$^3$LM$_{\mathtt{large(160G)}}$ and above strong baselines.
P$^3$LM$_{\mathtt{large(160G)}}$ outperforms all these published methods on GLGE according to the overall score.
Specifically, compared with the score 36.5 of ProphetNet$_\mathtt{large}$ which is the state-of-the-art published method, the score of our proposed P$^3$LM$_{\mathtt{large(160G)}}$ is 37.4, which achieves 0.9 absolute and 2.5\% relative improvements.
From the perspective of different tasks, the average scores of our model are 34.9/29.2/75.3/25.9, which are 34.5/28.5/73.0/23.6 for ProphetNet$_\mathtt{large}$, on text summarization, question generation, question answering, and persona dialog response generation, respectively. Our model achieves +0.6/+0.7/+2.3/+2.3 absolute improvements.
Based on the above results, the effectiveness of P$^3$LM is 
% comprehensively 
verified again.
Besides, L2R inference is explained in Appendix~\ref{sec:appendix_inference} and the effect of pre-training iterations is shown in Appendix~\ref{sec:appendix_iter}.

\begin{figure}[t]
	\centering
	\includegraphics[width=3.0in]{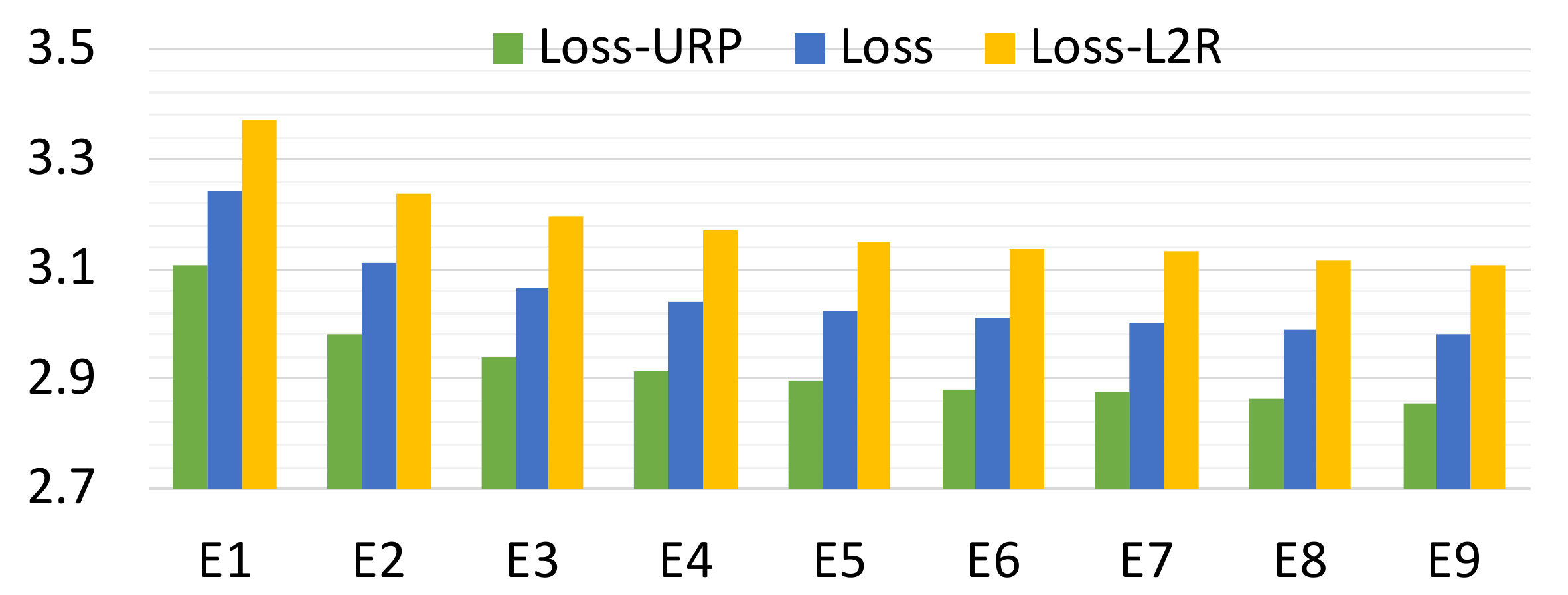}
	\caption{\label{loss} Losses of each epoch during the pre-training of P$^3$LM$_{\mathtt{large(160G)}}$. \textbf{Same tend for perplexity}.}
% 	\vspace{-0.4cm}
\end{figure}
% \subsection{Discussion}
\subsubsection{Order Matters for Language Modeling}
To explore the effect of orders, we split the loss of $\alpha$-P$^3$LM into two parts, i.e., loss-URP and loss-L2R.
The first part corresponds to $\alpha p^{\mathtt{URP}}$ in $p^{\alpha}$, and the second corresponds to $(1-\alpha)p^{\mathtt{L2R}}$ in $p^{\alpha}$.
Figure~\ref{loss} shows that the loss-URP fits faster than loss-L2R.
Since the perplexity $ppl=2^{loss}$, we conclude that {\textbf{URP order achieves lower perplexity than L2R order}}, i.e., the difficulty of modeling natural language sentence in an L2R order is larger than the average level reflected by the URP order.
This observation indicates that sequence order matters for language modelling.
In future, we will consider to train P$^3$LM in orders considering syntactical information, e.g., a level-order traversal of the syntactic tree of a natural language sequence.

\subsection{Finetuning on Text Summarization.\label{Summarization}}
Abstractive text summarization as a typical NLG task, aims to generate a short and fluent summary of a long text document.
In this section, we finetune and evaluate the proposed P$^3$LM on a text summarization dataset CNN/DM introduced before.
\subsubsection{Experiment Settings}
% CNN/DM~\cite{hermann2015cnndm,see2017get} is a widely used text summarization corpus, which
% % In this paper, we follow most of previous pre-training work and use the non-anonymized version of the dataset.
% contains 287,113 article and summary pairs as training set, 13,368 pairs as validation set, and 11,490 pairs as test set.
A base and a large models on CNN/DM with batch size as 512 are finetuned, and max epochs are set as 25 and 15, respectively.
Adam optimizer is used to update the parameters of the model with a learning rate of 1e-4 and warm-up updates of 1,000.
Model with the best rouge score on the validation set is used for testing.
Although a URP order is used for training the P$^3$LM, we use beam search with an L2R order to generate summaries during inference.
Beam size is set as 5 for both the base and large models.
% and 7 for the large model.
The length of the target sequence is limited between 45 and 110 with a length penalty as 1.2.
\begin{table}[t]
	\centering
	\small
	\begin{tabular}{lccc}
		\hline
		{\bf Method} &\bf R-1  & \bf R-2 & \bf R-L\\ 
		\hline \hline
		\multicolumn{4}{c}{w/o pre-training} \\
		\hline
		LEAD-3 &  40.42 & 17.62 & 36.67\\
		PGNet & 36.44 & 15.66 & 33.42\\
		PGNet+Coverage & 39.53 & 17.28 & 36.38\\
		Bottom-Up & {41.22} & {18.68} & {38.34}\\ 
		\hline
		\multicolumn{4}{c}{w/ pre-training} \\ 
		\hline
		S2S-ELMo & 41.56 & 18.94 & 38.47\\ 
		BERT-SUMABS & 41.72 & 19.39 & 38.76\\
		BERT-SUMEXTABS  & 42.13 & 19.60 & 39.18\\	
		MASS  &42.12 & 19.50 & 39.01\\
		UniLM  & 43.33 & 20.21 & 40.51\\
		PALM  & 42.71 & 19.97 &39.71 \\
		ProphetNet$_{\mathtt{base(16G)}}$  & 42.52 & 19.78 & 39.59\\
		ProphetNet$_{\mathtt{large(16G)}}$  & \underline{43.68} & \underline{20.64} & \underline{40.72}\\ 	
		\cdashline{1-4}[1pt/2pt]
		
% 		P$^3$LM$_{\mathtt{base}}$ & 16GB &500k &1024& 42.82 &  19.88 &  39.81\\
		P$^3$LM$_{\mathtt{base(16G)}}$  & 42.90 &  19.98 &  39.93\\ %beam = 5
% 		P$^3$LM$_{\mathtt{base(16G)}}$  & 42.82 & 19.90 & 39.83 \\ %beam = 7
% 		P$^3$LM$_{\mathtt{large(16G)}}$ & 43.99 & 20.87 & 41.06 \\ %beam 7
		P$^3$LM$_{\mathtt{large(16G)}}$ & \bf 44.07 & \bf 20.82 & \bf 41.15\\ %beam = 5
% 		P$^3$LM$_{\mathtt{large(16G)}}$  & \bf 44.09 & \bf 20.90 &\bf 41.15 \\  %beam = 7
		\hline \hline
	\end{tabular}
	\caption{\label{cnndm_16g} Experiment results on CNN/DM. Pre-training corpus of all methods is less than 18GB. 
	Highest scores are in bold, and seconds are underlined. 
	The gains of P$^3$LM$_{\mathtt{base(16G)}}$ over ProphetNet$_{\mathtt{base(16G)}}$, and  P$^3$LM$_{\mathtt{large(16G)}}$ over ProphetNet$_{\mathtt{large(16G)}}$ are statistically significant at $p = 0.05$.}
\end{table}
% \vspace{-0.2cm}
\subsubsection{Baselines}
Popular baselines are compared for evaluation.
% \begin{itemize}
% \item 
{LEAD-3}~\cite{nallapati2017summarunner} takes the first three sentences as the summary;
% \item 
{PGNet}~\cite{see2017get} is Seq2Seq model incorporated with a copy mechanism;
% \item 
{PGNet+Coverage}~\cite{see2017get} introduces a coverage mechanism to PGNet;
% \item 
{BottomUp}~\cite{gehrmann2018bottom} employs a bottom-up content selector based on Seq2Seq model;
% \item 
{S2S-ELMo}~\cite{edunov2019pre} uses the pre-trained ELMo~\cite{radford2018improving} representations for generation.
% \item 
Several pre-training based strong baselines including {BERTSUMABS}~\cite{liu2019text}, {MASS}, {UniLM}~\cite{dong2019unified}, {PALM}~\cite{bi2020palm}, and {ProphetNet} are also compared.
% \end{itemize}
% Specifically, ProphetNet is a state-of-the-art generative pre-training model incorporating prophet decoding mechanism.

%############################# Experiment Results #############################
\subsubsection{Experiment Results}
Table \ref{cnndm_16g} shows experiment results of models without pre-training or pre-trained on the less than 18GB wikipedia and bookcorpus dataset, where ELMo is an exception that it is trained on a 5GB dataset.
Results show that P$^3$LM outperforms the baselines and achieves the best performance.
Specifically, our base and large models achieve +0.38/+0.20/+0.34 and +0.39/+0.18/+0.43 improvements compared with corresponding ProphetNet models in terms of R-1, R-2, and R-L.
% The results indicate the effectiveness of the proposed P$^3$LM.
We think the improvements come from the P$^3$LM decoding that strengthens bi-direction information and long dependencies modeling of target sequences.
\begin{table}[t]
	\centering
	\small
% 	\resizebox{0.49\textwidth}{!}{
	\begin{tabular}{m{0.07\textwidth}m{0.15\textwidth}ccc}
		\hline 
		{\bf Init} &\bf Settings & \bf R-1  & \bf R-2 & \bf R-L\\ 
		\hline \hline
		\multicolumn{5}{c}{w/o pre-training} \\
		\hline
		
		\multirow{5}{*}{{base}}  
		& $N\!=\!1$, $p^{\mathtt{L2R}}$ & 40.33 & 17.64 & 37.35 \\
		& $N\!=\!2$, $p^{\mathtt{L2R}}$ & 40.65 & 17.94 & 37.67 \\ 
		& $N\!=\!2$, $p^{\mathtt{URP}}$ & 36.59 & 15.57 & 34.42 \\
% 		& L2R & 40.65 & 17.94 & 37.67 \\
		& $N\!=\!2$, $p^{\mathtt{URP}}\!\to\! p^{\mathtt{L2R}}$ & 41.32 & 18.57 & 38.46 \\
		& $N\!=\!2$, $p^{\alpha}$ & \bf 41.38 & \bf 18.60 & \bf 38.47\\ 
		\cdashline{1-5}[1pt/2pt]
		
		\multirow{5}{*}{{large}}
		& $N\!=\!1$, $p^{\mathtt{L2R}}$ & 40.53  & 17.54 & 37.61 \\
		& $N\!=\!2$, $p^{\mathtt{L2R}}$ & 41.04 & 18.12 & 38.08 \\
		& $N\!=\!2$, $p^{\mathtt{URP}}$ & 36.19 & 15.43 & 33.84 \\
% 		& L2R & 41.04 & 18.12 & 38.08 \\
		& $N\!=\!2$, $p^{\mathtt{URP}}\!\to\! p^{\mathtt{L2R}}$ & \bf 42.00 & \bf 19.02 & \bf 39.06 \\
		& $N\!=\!2$, $p^{\alpha}$ & 41.45 & 18.66 & 38.42\\ 
		\hline
		\multicolumn{5}{c}{w/ pre-training}  \\
% 		\multicolumn{5}{c}{w/ pre-trained P$^3$LM$_{\mathtt{base(16G)}}$ and P$^3$LM$_{\mathtt{large(16G)}}$} \\
		\hline
		\multirow{5}{*}{base$\mathtt{(16G)}$} 
		& $N\!=\!1$, $p^{\mathtt{L2R}}$ & 42.32 & 19.45 & 39.27\\
		& $N\!=\!2$, $p^{\mathtt{L2R}}$ & 42.56 & 19.57 & 39.56 \\
		& $N\!=\!2$, $p^{\mathtt{URP}}$ & 38.35 & 17.00 & 36.19 \\
% 		& L2R & 42.56 & 19.57 & 39.56\\
		& $N\!=\!2$, $p^{\mathtt{URP}}\!\to\! p^{\mathtt{L2R}}$ & \bf 42.92 & \bf 19.98 & \bf 39.93  \\
		& $N\!=\!2$,$p^{\alpha}$ & 42.90 & \bf 19.98 & \bf 39.93\\
		\cdashline{1-5}[1pt/2pt]
		
		\multirow{5}{*}{large$\mathtt{(16G)}$}
		& $N\!=\!1$, $p^{\mathtt{L2R}}$ & 43.27 & 20.10 & 40.23 \\
		& $N\!=\!2$, $p^{\mathtt{L2R}}$ & 43.43 & 20.18 & 40.46 \\
		& $N\!=\!2$, $p^{\mathtt{URP}}$ & 38.94 & 17.61 & 36.76 \\
% 		& L2R & 43.43 & 20.18 & 40.46\\
		& $N\!=\!2$, $p^{\mathtt{URP}}\!\to\! p^{\mathtt{L2R}}$ & 43.60 & 20.59 & 40.66 \\
		& $N\!=\!2$, $p^{\alpha}$ & \bf 44.07 & \bf 20.82 & \bf 41.15\\
		\hline \hline
%       P$^3$LM$_{\mathtt{base}}$ w/ p(125k\_best) & URP & 38.26 & 17.01 & 36.06 \\
% 		P$^3$LM$_{\mathtt{base}}$ w/ p(125k\_best) & URP$\to$L2R & 42.35 & 19.25 & 39.37 \\
% 		P$^3$LM$_{\mathtt{base}}$ w/ p(500k\_last) & URP & 38.77 & 17.21& 36.52 \\
% 		P$^3$LM$_{\mathtt{base}}$ w/ p(500k\_last) & L2R & 42.49 & 19.58 & 39.44\\
% 		P$^3$LM$_{\mathtt{base}}$ w/ p(500k\_last) & URP$\to$L2R & 42.60 & 19.59 & 39.77 \\
% 		P$^3$LM$_{\mathtt{base}}$ w/ p(500k\_last) & L2R$+$URP & \bf 42.82 & \bf 19.88 & \bf 39.81\\ \hline
	\end{tabular}
% 	}
	\caption{\label{finetune_objs_results} Ablation study of different finetuning settings on CNN/DM, with pre-training or not.}
% 	Highest score are in bold. }
% 	\textbf{Rnd} means the models are randomly initialized according to a uniform distribution.
\vspace{-0.2cm}
\end{table}
% \vspace{-0.1cm}
\subsubsection{Ablation Study\label{Objectives}}
To further verify the effectiveness of the proposed P$^3$LM, we conduct ablation study on different finetuning settings.
We investigate different combinations of finetuning settings and show the results in Table~\ref{finetune_objs_results}.
% including $\mathtt{L2R}$, $\mathtt{L2R}_p$, $\mathtt{URP}_p$, $\mathtt{URP}_p\to\mathtt{L2R}_p$, and $\mathtt{URP}_p+\mathtt{L2R}_p$.
% Specifically, compared with $\mathtt{L2R}$, $\mathtt{L2R}_p$ has an additional objective which predicts one word ahead.
Specifically, $p^{\mathtt{URP}}\!\to\! p^{\mathtt{L2R}}$ means the model is firstly trained several epochs on sampled instances with orders subjecting to distribution $p^{\mathtt{URP}}$ and then several epochs to distribution $p^{\mathtt{L2R}}$.
% $\mathtt{URP}_p+\mathtt{L2R}_p$ means the model is trained with both of the two objectives at the same time.
% , where ``Rnd" means the parameters are randomly initialized according to a uniform distribution.
% Tns-XL is short for transformer-XL~\cite{DBLP:conf/acl/DaiYYCLS19}, which is an multi-stream version of the transformer~\cite{transformer}.
% Both P$^3$LM and ProphetNet has the same neural architecture as Tns-XL.
We first observe that prophet mechanism ($N\!=\!2$) brings improvements.
More importantly, compared with $p^{\mathtt{L2R}}$, P$^3$LM introduces $p^{\mathtt{URP}}$, where we can see that the $p^{\mathtt{URP}}\!\to\!p^{\mathtt{L2R}}$ and $p^{\alpha}$ achieve the best performance when loading the pre-trained P$^3$LM$_{\mathtt{base(16G)}}$ and P$^3$LM$_{\mathtt{large(16G)}}$ models.
Furthermore, when loading no pre-trained models, P$^3$LM trained based on $p^{\mathtt{URP}}\!\to\!p^{\mathtt{L2R}}$ and $p^{\alpha}$ still improve traditional $\mathtt{L2R}$ training a lot.
P$^3$LM with only $p^{\mathtt{URP}}$ performs the worst, which is reasonable since the model only uniformly selects one permutation of a target as training data, which is completely inconsistent with the L2R inference. It further indicates that, although the L2R order is only one special case of all $T!$ permutations, it is still important and should be paid more attention as our $\alpha$-P$^3$LM do.

\section{Conclusion}
A probabilistically permuted prophet language modeling, P$^3$LM, is proposed for generative pre-training.
P$^3$LM models sequences by considering both left-to-right and random permutation orders, equipped with a prophet mechanism for future token prediction.
Extensive experiments are conducted on GLGE, a general natural language generation evaluation benchmark,
% on summarization, question generation, and machine reading comprehension tasks, 
where P$^3$LM achieves state-of-the-art results compared with public available generative pre-training methods.

\section*{Limitations}
\subsection*{Exploring Better Distribution $p^{*}(Z)$}
Figure~\ref{loss} shows that URP loss fits faster than L2R loss.
Since L2R is a special case of URP order, we think that the difficulty of modeling natural language sentence in an L2R order is larger than the average level reflected by the URP order.
It indicates that sequence order really matters for language modelling and exploring other distribution $p^{*}(Z)$ besides $p^{\alpha}$ could be an interesting problem. 
In future, we will consider to train P$^3$LM in orders considering syntax, e.g., a level-order traversal of the syntactic tree of a natural language.

\subsection*{Training-Inference Consistency}
\label{sec:appendix_inference}
P$^3$LM decodes a sequence in an order sampled from $p^{\alpha}$ during training.
Different from training, P$^3$LM performs L2R decoding during inference.
Nevertheless, P$^3$LM achieves significant improvements across multiple tasks and datasets.
We think this benefits from the involving of P$^3$LM decoding which introduces more constraints to help the model to learn bidirectional context and long dependency modeling.
In future, we will explore to decode a sequence in terms of an optimized order, not limited in L2R order.
% , to further improve the effectiveness of our model.

\subsection*{Training Efficiency.}
The model construction and network structure is as complex as ProphetNet. The key point of P$^3$LM is utilizing sampled orders according to a given distribution as the attention mask in transformer decoder.
This makes the computation cost of P$^3$LM similar to ProphetNet when sampling one order for a target sequence.
In this paper, according to experiments, we sample two orders from $p^\alpha$ for training, this makes training one instance in one epoch twice the time of ProphetNet. 
% However, the total epochs for training could be reduced for saving time.

\section*{Acknowledge}
Supported by the National Key R\&D Program of China under Grant No. 2020AAA0108600.

\bibliographystyle{acl_natbib}
\bibliography{p3lm}

\begin{thebibliography}{40}
\expandafter\ifx\csname natexlab\endcsname\relax\def\natexlab#1{#1}\fi

\bibitem[{an(2019)}]{DBLP:journals/corr/abs-1905-12790}
Elman~Mansimov an. 2019.
\newblock \href {https://arxiv.org/abs/1905.12790} {A generalized framework of
  sequence generation with application t}.
\newblock \emph{ArXiv preprint}, abs/1905.12790.

\bibitem[{Bahdanau et~al.(2015)Bahdanau, Cho, and Bengio}]{bahdanau2014neural}
Dzmitry Bahdanau, Kyunghyun Cho, and Yoshua Bengio. 2015.
\newblock \href {http://arxiv.org/abs/1409.0473} {Neural machine translation by
  jointly learning to align and translate}.
\newblock In \emph{3rd International Conference on Learning Representations,
  {ICLR} 2015, San Diego, CA, USA, May 7-9, 2015, Conference Track
  Proceedings}.

\bibitem[{Banerjee and Lavie(2005)}]{banerjee2005meteor}
Satanjeev Banerjee and Alon Lavie. 2005.
\newblock \href {https://aclanthology.org/W05-0909} {{METEOR}: An automatic
  metric for {MT} evaluation with improved correlation with human judgments}.
\newblock In \emph{Proceedings of the {ACL} Workshop on Intrinsic and Extrinsic
  Evaluation Measures for Machine Translation and/or Summarization}, pages
  65--72, Ann Arbor, Michigan. Association for Computational Linguistics.

\bibitem[{Bi et~al.(2020)Bi, Li, Wu, Yan, Wang, Huang, Huang, and
  Si}]{bi2020palm}
Bin Bi, Chenliang Li, Chen Wu, Ming Yan, Wei Wang, Songfang Huang, Fei Huang,
  and Luo Si. 2020.
\newblock \href {https://doi.org/10.18653/v1/2020.emnlp-main.700} {{PALM}:
  Pre-training an autoencoding{\&}autoregressive language model for
  context-conditioned generation}.
\newblock In \emph{Proceedings of the 2020 Conference on Empirical Methods in
  Natural Language Processing (EMNLP)}, pages 8681--8691, Online. Association
  for Computational Linguistics.

\bibitem[{Brown et~al.(2020)Brown, Mann, Ryder, Subbiah, Kaplan, Dhariwal,
  Neelakantan, Shyam, Sastry, Askell, Agarwal, Herbert{-}Voss, Krueger,
  Henighan, Child, Ramesh, Ziegler, Wu, Winter, Hesse, Chen, Sigler, Litwin,
  Gray, Chess, Clark, Berner, McCandlish, Radford, Sutskever, and
  Amodei}]{brown2020language}
Tom~B. Brown, Benjamin Mann, Nick Ryder, Melanie Subbiah, Jared Kaplan,
  Prafulla Dhariwal, Arvind Neelakantan, Pranav Shyam, Girish Sastry, Amanda
  Askell, Sandhini Agarwal, Ariel Herbert{-}Voss, Gretchen Krueger, Tom
  Henighan, Rewon Child, Aditya Ramesh, Daniel~M. Ziegler, Jeffrey Wu, Clemens
  Winter, Christopher Hesse, Mark Chen, Eric Sigler, Mateusz Litwin, Scott
  Gray, Benjamin Chess, Jack Clark, Christopher Berner, Sam McCandlish, Alec
  Radford, Ilya Sutskever, and Dario Amodei. 2020.
\newblock \href
  {https://proceedings.neurips.cc/paper/2020/hash/1457c0d6bfcb4967418bfb8ac142f64a-Abstract.html}
  {Language models are few-shot learners}.
\newblock In \emph{Advances in Neural Information Processing Systems 33: Annual
  Conference on Neural Information Processing Systems 2020, NeurIPS 2020,
  December 6-12, 2020, virtual}.

\bibitem[{Clark et~al.(2020)Clark, Luong, Le, and Manning}]{clark2020electra}
Kevin Clark, Minh{-}Thang Luong, Quoc~V. Le, and Christopher~D. Manning. 2020.
\newblock \href {https://openreview.net/forum?id=r1xMH1BtvB} {{ELECTRA:}
  pre-training text encoders as discriminators rather than generators}.
\newblock In \emph{8th International Conference on Learning Representations,
  {ICLR} 2020, Addis Ababa, Ethiopia, April 26-30, 2020}. OpenReview.net.

\bibitem[{Devlin et~al.(2019)Devlin, Chang, Lee, and
  Toutanova}]{devlin2018bert}
Jacob Devlin, Ming-Wei Chang, Kenton Lee, and Kristina Toutanova. 2019.
\newblock \href {https://doi.org/10.18653/v1/N19-1423} {{BERT}: Pre-training of
  deep bidirectional transformers for language understanding}.
\newblock In \emph{Proceedings of the 2019 Conference of the North {A}merican
  Chapter of the Association for Computational Linguistics: Human Language
  Technologies, Volume 1 (Long and Short Papers)}, pages 4171--4186,
  Minneapolis, Minnesota. Association for Computational Linguistics.

\bibitem[{Dong et~al.(2019)Dong, Yang, Wang, Wei, Liu, Wang, Gao, Zhou, and
  Hon}]{dong2019unified}
Li~Dong, Nan Yang, Wenhui Wang, Furu Wei, Xiaodong Liu, Yu~Wang, Jianfeng Gao,
  Ming Zhou, and Hsiao{-}Wuen Hon. 2019.
\newblock \href
  {https://proceedings.neurips.cc/paper/2019/hash/c20bb2d9a50d5ac1f713f8b34d9aac5a-Abstract.html}
  {Unified language model pre-training for natural language understanding and
  generation}.
\newblock In \emph{Advances in Neural Information Processing Systems 32: Annual
  Conference on Neural Information Processing Systems 2019, NeurIPS 2019,
  December 8-14, 2019, Vancouver, BC, Canada}, pages 13042--13054.

\bibitem[{Edunov et~al.(2019)Edunov, Baevski, and Auli}]{edunov2019pre}
Sergey Edunov, Alexei Baevski, and Michael Auli. 2019.
\newblock \href {https://doi.org/10.18653/v1/N19-1409} {Pre-trained language
  model representations for language generation}.
\newblock In \emph{Proceedings of the 2019 Conference of the North {A}merican
  Chapter of the Association for Computational Linguistics: Human Language
  Technologies, Volume 1 (Long and Short Papers)}, pages 4052--4059,
  Minneapolis, Minnesota. Association for Computational Linguistics.

\bibitem[{Emelianenko et~al.(2019)Emelianenko, Voita, and
  Serdyukov}]{DBLP:conf/nips/EmelianenkoVS19}
Dmitrii Emelianenko, Elena Voita, and Pavel Serdyukov. 2019.
\newblock \href
  {https://proceedings.neurips.cc/paper/2019/hash/1558417b096b5d8e7cbe0183ea9cbf26-Abstract.html}
  {Sequence modeling with unconstrained generation order}.
\newblock In \emph{Advances in Neural Information Processing Systems 32: Annual
  Conference on Neural Information Processing Systems 2019, NeurIPS 2019,
  December 8-14, 2019, Vancouver, BC, Canada}, pages 7698--7709.

\bibitem[{Gehrmann et~al.(2018)Gehrmann, Deng, and Rush}]{gehrmann2018bottom}
Sebastian Gehrmann, Yuntian Deng, and Alexander Rush. 2018.
\newblock \href {https://doi.org/10.18653/v1/D18-1443} {Bottom-up abstractive
  summarization}.
\newblock In \emph{Proceedings of the 2018 Conference on Empirical Methods in
  Natural Language Processing}, pages 4098--4109, Brussels, Belgium.
  Association for Computational Linguistics.

\bibitem[{Graff et~al.(2003)Graff, Kong, Chen, and Maeda}]{graff2003gigaword}
David Graff, Junbo Kong, Ke~Chen, and Kazuaki Maeda. 2003.
\newblock English gigaword.
\newblock \emph{Linguistic Data Consortium, Philadelphia}, 4(1):34.

\bibitem[{Gu et~al.(2019)Gu, Liu, and Cho}]{DBLP:journals/tacl/GuLC19}
Jiatao Gu, Qi~Liu, and Kyunghyun Cho. 2019.
\newblock \href {https://doi.org/10.1162/tacl_a_00292} {Insertion-based
  decoding with automatically inferred generation order}.
\newblock \emph{Transactions of the Association for Computational Linguistics},
  7:661--676.

\bibitem[{Hermann et~al.(2015)Hermann, Kocisk{\'{y}}, Grefenstette, Espeholt,
  Kay, Suleyman, and Blunsom}]{hermann2015cnndm}
Karl~Moritz Hermann, Tom{\'{a}}s Kocisk{\'{y}}, Edward Grefenstette, Lasse
  Espeholt, Will Kay, Mustafa Suleyman, and Phil Blunsom. 2015.
\newblock \href
  {https://proceedings.neurips.cc/paper/2015/hash/afdec7005cc9f14302cd0474fd0f3c96-Abstract.html}
  {Teaching machines to read and comprehend}.
\newblock In \emph{Advances in Neural Information Processing Systems 28: Annual
  Conference on Neural Information Processing Systems 2015, December 7-12,
  2015, Montreal, Quebec, Canada}, pages 1693--1701.

\bibitem[{Kingma and Ba(2015)}]{kingma2014adam}
Diederik~P. Kingma and Jimmy Ba. 2015.
\newblock \href {http://arxiv.org/abs/1412.6980} {Adam: {A} method for
  stochastic optimization}.
\newblock In \emph{3rd International Conference on Learning Representations,
  {ICLR} 2015, San Diego, CA, USA, May 7-9, 2015, Conference Track
  Proceedings}.

\bibitem[{Lan et~al.(2020)Lan, Chen, Goodman, Gimpel, Sharma, and
  Soricut}]{lan2019albert}
Zhenzhong Lan, Mingda Chen, Sebastian Goodman, Kevin Gimpel, Piyush Sharma, and
  Radu Soricut. 2020.
\newblock \href {https://openreview.net/forum?id=H1eA7AEtvS} {{ALBERT:} {A}
  lite {BERT} for self-supervised learning of language representations}.
\newblock In \emph{8th International Conference on Learning Representations,
  {ICLR} 2020, Addis Ababa, Ethiopia, April 26-30, 2020}. OpenReview.net.

\bibitem[{Lewis et~al.(2020)Lewis, Liu, Goyal, Ghazvininejad, Mohamed, Levy,
  Stoyanov, and Zettlemoyer}]{lewis2019bart}
Mike Lewis, Yinhan Liu, Naman Goyal, Marjan Ghazvininejad, Abdelrahman Mohamed,
  Omer Levy, Veselin Stoyanov, and Luke Zettlemoyer. 2020.
\newblock \href {https://doi.org/10.18653/v1/2020.acl-main.703} {{BART}:
  Denoising sequence-to-sequence pre-training for natural language generation,
  translation, and comprehension}.
\newblock In \emph{Proceedings of the 58th Annual Meeting of the Association
  for Computational Linguistics}, pages 7871--7880, Online. Association for
  Computational Linguistics.

\bibitem[{Li et~al.(2016)Li, Galley, Brockett, Gao, and
  Dolan}]{li2015diversity}
Jiwei Li, Michel Galley, Chris Brockett, Jianfeng Gao, and Bill Dolan. 2016.
\newblock \href {https://doi.org/10.18653/v1/N16-1014} {A diversity-promoting
  objective function for neural conversation models}.
\newblock In \emph{Proceedings of the 2016 Conference of the North {A}merican
  Chapter of the Association for Computational Linguistics: Human Language
  Technologies}, pages 110--119, San Diego, California. Association for
  Computational Linguistics.

\bibitem[{Lin(2004)}]{lin2004rouge}
Chin-Yew Lin. 2004.
\newblock \href {https://aclanthology.org/W04-1013} {{ROUGE}: A package for
  automatic evaluation of summaries}.
\newblock In \emph{Text Summarization Branches Out}, pages 74--81, Barcelona,
  Spain. Association for Computational Linguistics.

\bibitem[{Liu et~al.(2021)Liu, Yan, Gong, Qi, Zhang, Jiao, Chen, Fu, Shou,
  Gong, Wang, Chen, Jiang, Lv, Zhang, Wu, Zhou, and Duan}]{Liu2020GLGE}
Dayiheng Liu, Yu~Yan, Yeyun Gong, Weizhen Qi, Hang Zhang, Jian Jiao, Weizhu
  Chen, Jie Fu, Linjun Shou, Ming Gong, Pengcheng Wang, Jiusheng Chen, Daxin
  Jiang, Jiancheng Lv, Ruofei Zhang, Winnie Wu, Ming Zhou, and Nan Duan. 2021.
\newblock \href {https://doi.org/10.18653/v1/2021.findings-acl.36} {{GLGE}: A
  new general language generation evaluation benchmark}.
\newblock In \emph{Findings of the Association for Computational Linguistics:
  ACL-IJCNLP 2021}, pages 408--420, Online. Association for Computational
  Linguistics.

\bibitem[{Liu and Lapata(2019)}]{liu2019text}
Yang Liu and Mirella Lapata. 2019.
\newblock \href {https://doi.org/10.18653/v1/D19-1387} {Text summarization with
  pretrained encoders}.
\newblock In \emph{Proceedings of the 2019 Conference on Empirical Methods in
  Natural Language Processing and the 9th International Joint Conference on
  Natural Language Processing (EMNLP-IJCNLP)}, pages 3730--3740, Hong Kong,
  China. Association for Computational Linguistics.

\bibitem[{Liu et~al.(2019)Liu, Ott, Goyal, Du, Joshi, Chen, Levy, Lewis,
  Zettlemoyer, and Stoyanov}]{liu2019roberta}
Yinhan Liu, Myle Ott, Naman Goyal, Jingfei Du, Mandar Joshi, Danqi Chen, Omer
  Levy, Mike Lewis, Luke Zettlemoyer, and Veselin Stoyanov. 2019.
\newblock \href {https://arxiv.org/abs/1907.11692} {Roberta: A robustly
  optimized bert pretraining approach}.
\newblock \emph{ArXiv preprint}, abs/1907.11692.

\bibitem[{Mikolov et~al.(2013)Mikolov, Chen, Corrado, and
  Dean}]{mikolov2013efficient}
Tomas Mikolov, Kai Chen, Greg Corrado, and Jeffrey Dean. 2013.
\newblock \href {https://arxiv.org/abs/1301.3781} {Efficient estimation of word
  representations in vector space}.
\newblock \emph{ArXiv preprint}, abs/1301.3781.

\bibitem[{Nallapati et~al.(2017)Nallapati, Zhai, and
  Zhou}]{nallapati2017summarunner}
Ramesh Nallapati, Feifei Zhai, and Bowen Zhou. 2017.
\newblock \href {http://aaai.org/ocs/index.php/AAAI/AAAI17/paper/view/14636}
  {Summarunner: {A} recurrent neural network based sequence model for
  extractive summarization of documents}.
\newblock In \emph{Proceedings of the Thirty-First {AAAI} Conference on
  Artificial Intelligence, February 4-9, 2017, San Francisco, California,
  {USA}}, pages 3075--3081. {AAAI} Press.

\bibitem[{Narayan et~al.(2018)Narayan, Cohen, and Lapata}]{narayan2018don}
Shashi Narayan, Shay~B. Cohen, and Mirella Lapata. 2018.
\newblock \href {https://doi.org/10.18653/v1/D18-1206} {Don{'}t give me the
  details, just the summary! topic-aware convolutional neural networks for
  extreme summarization}.
\newblock In \emph{Proceedings of the 2018 Conference on Empirical Methods in
  Natural Language Processing}, pages 1797--1807, Brussels, Belgium.
  Association for Computational Linguistics.

\bibitem[{Papineni et~al.(2002)Papineni, Roukos, Ward, and
  Zhu}]{papineni2002bleu}
Kishore Papineni, Salim Roukos, Todd Ward, and Wei-Jing Zhu. 2002.
\newblock \href {https://doi.org/10.3115/1073083.1073135} {{B}leu: a method for
  automatic evaluation of machine translation}.
\newblock In \emph{Proceedings of the 40th Annual Meeting of the Association
  for Computational Linguistics}, pages 311--318, Philadelphia, Pennsylvania,
  USA. Association for Computational Linguistics.

\bibitem[{Qi et~al.(2020)Qi, Yan, Gong, Liu, Duan, Chen, Zhang, and
  Zhou}]{yan2020prophetnet}
Weizhen Qi, Yu~Yan, Yeyun Gong, Dayiheng Liu, Nan Duan, Jiusheng Chen, Ruofei
  Zhang, and Ming Zhou. 2020.
\newblock \href {https://doi.org/10.18653/v1/2020.findings-emnlp.217}
  {{P}rophet{N}et: Predicting future n-gram for
  sequence-to-{S}equence{P}re-training}.
\newblock In \emph{Findings of the Association for Computational Linguistics:
  EMNLP 2020}, pages 2401--2410, Online. Association for Computational
  Linguistics.

\bibitem[{Radford et~al.()Radford, Narasimhan, Salimans, and
  Sutskever}]{radford2018improving}
Alec Radford, Karthik Narasimhan, Tim Salimans, and Ilya Sutskever.
\newblock Improving language understanding by generative pre-training.

\bibitem[{Raffel et~al.(2019)Raffel, Shazeer, Roberts, Lee, Narang, Matena,
  Zhou, Li, and Liu}]{raffel2019exploring}
Colin Raffel, Noam Shazeer, Adam Roberts, Katherine Lee, Sharan Narang, Michael
  Matena, Yanqi Zhou, Wei Li, and Peter~J Liu. 2019.
\newblock \href {https://arxiv.org/abs/1910.10683} {Exploring the limits of
  transfer learning with a unified text-to-text transformer}.
\newblock \emph{ArXiv preprint}, abs/1910.10683.

\bibitem[{Rajpurkar et~al.(2016)Rajpurkar, Zhang, Lopyrev, and
  Liang}]{rajpurkar2016squad}
Pranav Rajpurkar, Jian Zhang, Konstantin Lopyrev, and Percy Liang. 2016.
\newblock \href {https://doi.org/10.18653/v1/D16-1264} {{SQ}u{AD}: 100,000+
  questions for machine comprehension of text}.
\newblock In \emph{Proceedings of the 2016 Conference on Empirical Methods in
  Natural Language Processing}, pages 2383--2392, Austin, Texas. Association
  for Computational Linguistics.

\bibitem[{Reddy et~al.(2019)Reddy, Chen, and Manning}]{reddy2019coqa}
Siva Reddy, Danqi Chen, and Christopher~D. Manning. 2019.
\newblock \href {https://doi.org/10.1162/tacl_a_00266} {{C}o{QA}: A
  conversational question answering challenge}.
\newblock \emph{Transactions of the Association for Computational Linguistics},
  7:249--266.

\bibitem[{Rush et~al.(2015)Rush, Chopra, and Weston}]{rush2015neural}
Alexander~M. Rush, Sumit Chopra, and Jason Weston. 2015.
\newblock \href {https://doi.org/10.18653/v1/D15-1044} {A neural attention
  model for abstractive sentence summarization}.
\newblock In \emph{Proceedings of the 2015 Conference on Empirical Methods in
  Natural Language Processing}, pages 379--389, Lisbon, Portugal. Association
  for Computational Linguistics.

\bibitem[{See et~al.(2017)See, Liu, and Manning}]{see2017get}
Abigail See, Peter~J. Liu, and Christopher~D. Manning. 2017.
\newblock \href {https://doi.org/10.18653/v1/P17-1099} {Get to the point:
  Summarization with pointer-generator networks}.
\newblock In \emph{Proceedings of the 55th Annual Meeting of the Association
  for Computational Linguistics (Volume 1: Long Papers)}, pages 1073--1083,
  Vancouver, Canada. Association for Computational Linguistics.

\bibitem[{Song et~al.(2019)Song, Tan, Qin, Lu, and Liu}]{song2019mass}
Kaitao Song, Xu~Tan, Tao Qin, Jianfeng Lu, and Tie{-}Yan Liu. 2019.
\newblock \href {http://proceedings.mlr.press/v97/song19d.html} {{MASS:} masked
  sequence to sequence pre-training for language generation}.
\newblock In \emph{Proceedings of the 36th International Conference on Machine
  Learning, {ICML} 2019, 9-15 June 2019, Long Beach, California, {USA}},
  volume~97 of \emph{Proceedings of Machine Learning Research}, pages
  5926--5936. {PMLR}.

\bibitem[{Vaswani et~al.(2017)Vaswani, Shazeer, Parmar, Uszkoreit, Jones,
  Gomez, Kaiser, and Polosukhin}]{transformer}
Ashish Vaswani, Noam Shazeer, Niki Parmar, Jakob Uszkoreit, Llion Jones,
  Aidan~N. Gomez, Lukasz Kaiser, and Illia Polosukhin. 2017.
\newblock \href
  {https://proceedings.neurips.cc/paper/2017/hash/3f5ee243547dee91fbd053c1c4a845aa-Abstract.html}
  {Attention is all you need}.
\newblock In \emph{Advances in Neural Information Processing Systems 30: Annual
  Conference on Neural Information Processing Systems 2017, December 4-9, 2017,
  Long Beach, CA, {USA}}, pages 5998--6008.

\bibitem[{Vinyals et~al.(2016)Vinyals, Bengio, and
  Kudlur}]{DBLP:journals/corr/VinyalsBK15}
Oriol Vinyals, Samy Bengio, and Manjunath Kudlur. 2016.
\newblock \href {http://arxiv.org/abs/1511.06391} {Order matters: Sequence to
  sequence for sets}.
\newblock In \emph{4th International Conference on Learning Representations,
  {ICLR} 2016, San Juan, Puerto Rico, May 2-4, 2016, Conference Track
  Proceedings}.

\bibitem[{Xiao et~al.(2020)Xiao, Zhang, Li, Sun, Tian, Wu, and
  Wang}]{xiao2020ernie}
Dongling Xiao, Han Zhang, Yu{-}Kun Li, Yu~Sun, Hao Tian, Hua Wu, and Haifeng
  Wang. 2020.
\newblock \href {https://doi.org/10.24963/ijcai.2020/553} {{ERNIE-GEN:} an
  enhanced multi-flow pre-training and fine-tuning framework for natural
  language generation}.
\newblock In \emph{Proceedings of the Twenty-Ninth International Joint
  Conference on Artificial Intelligence, {IJCAI} 2020}, pages 3997--4003.
  ijcai.org.

\bibitem[{Yang et~al.(2019)Yang, Dai, Yang, Carbonell, Salakhutdinov, and
  Le}]{yang2019xlnet}
Zhilin Yang, Zihang Dai, Yiming Yang, Jaime~G. Carbonell, Ruslan Salakhutdinov,
  and Quoc~V. Le. 2019.
\newblock \href
  {https://proceedings.neurips.cc/paper/2019/hash/dc6a7e655d7e5840e66733e9ee67cc69-Abstract.html}
  {Xlnet: Generalized autoregressive pretraining for language understanding}.
\newblock In \emph{Advances in Neural Information Processing Systems 32: Annual
  Conference on Neural Information Processing Systems 2019, NeurIPS 2019,
  December 8-14, 2019, Vancouver, BC, Canada}, pages 5754--5764.

\bibitem[{Zhang et~al.(2019)Zhang, Zhao, Saleh, and Liu}]{zhang2019pegasus}
Jingqing Zhang, Yao Zhao, Mohammad Saleh, and Peter~J Liu. 2019.
\newblock \href {https://arxiv.org/abs/1912.08777} {Pegasus: Pre-training with
  extracted gap-sentences for abstractive summarization}.
\newblock \emph{ArXiv preprint}, abs/1912.08777.

\bibitem[{Zhang et~al.(2018)Zhang, Dinan, Urbanek, Szlam, Kiela, and
  Weston}]{zhang2018personalizing}
Saizheng Zhang, Emily Dinan, Jack Urbanek, Arthur Szlam, Douwe Kiela, and Jason
  Weston. 2018.
\newblock \href {https://doi.org/10.18653/v1/P18-1205} {Personalizing dialogue
  agents: {I} have a dog, do you have pets too?}
\newblock In \emph{Proceedings of the 56th Annual Meeting of the Association
  for Computational Linguistics (Volume 1: Long Papers)}, pages 2204--2213,
  Melbourne, Australia. Association for Computational Linguistics.

\end{thebibliography}
% \appendix
% \section{Impact Statement}
% aaaa

\clearpage

\appendix

\section{P$^3$LM v.s. XLNet}
\label{sec:appendix_differ}
The idea of permuted decoding is inspired by XLNet. However, P$^3$LM is different from XLNet in multiple aspects as follows. 
First, P$^3$LM is designed for addressing bi-directional context and long dependency problems for natural language generation (NLG), while XLNet is for natural language understanding (NLU);
Second, P$^3$LM is in a transformer encoder-decoder architecture, while XLNet is only a transformer encoder;
Third, P$^3$LMis trained on the full permutation of the {target} sequence to enhance long dependency modeling, while XLNet is trained on partial permutation of the {source} sequence; 
Fourth, P$^3$LM is implemented with multi-streams ($\#>=3$) for predicting multiple future tokens at one time step, while XLNET is implemented with two streams ($\#=2$) for predicting one token at a step;
Fifth, P$^3$LM implements permuted decoding that requiring a shift-right operation while XLNet does not, which is due to transformer's different designs for encoder and decoder.

\section{Model Parameters on GLGE}
\label{sec:appendix_param}
Table~\ref{P2DeNET_imp} shows the parameters of our model on GLGE.
Parameters are primarily searched from LR$\in$\{1e-4, 1e-5\}, WarmUp $\in$ \{0.5k, 1k\}, BatchSize $\in$ \{128, 256, 512\}, BeamSize $\in$ [4, 10], and LenPenalty $\in$ [0.6, 1.5], except WarmUp=10k for Gigaword and LenPenalty=10.0 for PersonaChat.
\begin{table}[h]
	\centering
	\small
% 	\scriptsize
    \resizebox{0.48\textwidth}{!}{
	\begin{tabular}{lcccccccc}
	\hline \hline
	\multirow{2}{*}{\bf Parameters} & \multicolumn{4}{c}{\bf Text Summarization} & \multicolumn{2}{c}{\bf QG}& \multicolumn{1}{c}{\bf QA}& \multicolumn{1}{c}{\bf Dialog} \\

% 	&{\bf CNN/DM} & {\bf Gigaword} & {\bf XSUM} & {\bf MSNews} & {\bf SQuAD} & {\bf MSQG} & {\bf CoQA} & {\bf PersonaChat} \\
	&{\bf CD} & {\bf GG} & {\bf XS} & {\bf MN} & {\bf SQ} & {\bf MQ} & {\bf CQ} & {\bf PC} \\
	\hline
	LR & 1e-4 & 1e-4 & 1e-4 & 1e-5 & 1e-5 & 1e-5 & 1e-5 & 1e-4\\
	WarmUp & 1k & 10k & 0.5k & 1k & 1k & 1k & 1k & 0.5k\\
	BatchSize & 512 & 128 & 256 & 128 & 128 & 128 & 128 & 128\\
	MaxEpoch & 15 & 6 & 15 & 15 & 10 & 10 & 10 & 15\\
	MaxSrcLen & 512 & 128 & 512 & 512 & 256 & 256 & 512 & 256\\
	MaxTgtLen & 128 & 32 & 128 & 64 & 64 & 32 & 32 & 32\\ \cdashline{1-9}[1pt/2pt]
	BeamSize & 5 & 5 & 8 & 8 & 6 & 4 & 7 & 10\\
	LenPenalty & 1.4 & 0.9 & 0.8 & 0.9 & 1.0 & 0.8 & 0.8 & 10.0\\
% 	DecMinLen & 45 & 3 & 10 & 3-64 & 5-32 & 3-32 & 1-32 & 3-32\\
	DecLen & 45-110 & 3-32 & 10-64 & 3-64 & 5-32 & 3-32 & 1-32 & 3-32\\
	BestEpoch & 14 & 6 & 9 & 13 & 7 & 5 & 10 & 13 \\
	\hline \hline
	\end{tabular}
	}
	\caption{\label{P2DeNET_imp} Hyperparameters used in fine-tuning P$^3$LM on GLGE. LR: learning rate. WarmUp: warm up steps. BatchSize: batch size. MaxEpoch: max epochs in fine-tuning. MaxSrcLen: source max length. MaxTgtLen:target max lengt. BeamSize: decoding beam size. LenPenalty: decoding length penalty. DecLen: length range of generated sequence. BestEpoch: best performing epoch. CD: CNN/DM. GG: Gigaword. XS: XSUM. MN: MSNews. SQ: SQuAD-QG. MQ: MSQG. CQ: CoQA. PC: PersonaChat.}
\end{table}

\section{L2R Inference}
\label{sec:appendix_inference}
P$^3$LM decodes a sequence in both L2R and URP order with prophet mechanism during training.
Different from training, our model leverages L2R decoding during inference.
Nevertheless, P$^3$LM achieves significant improvements across multiple tasks and datasets.
We think this benefits from the involving of P$^3$LM decoding which introduces more constraints to help the model to learn bidirectional context and long dependency modeling.

\section{Effect of Pre-training Iterations }
\label{sec:appendix_iter}
\begin{figure}[h]
	\centering
	\includegraphics[width=3.0in]{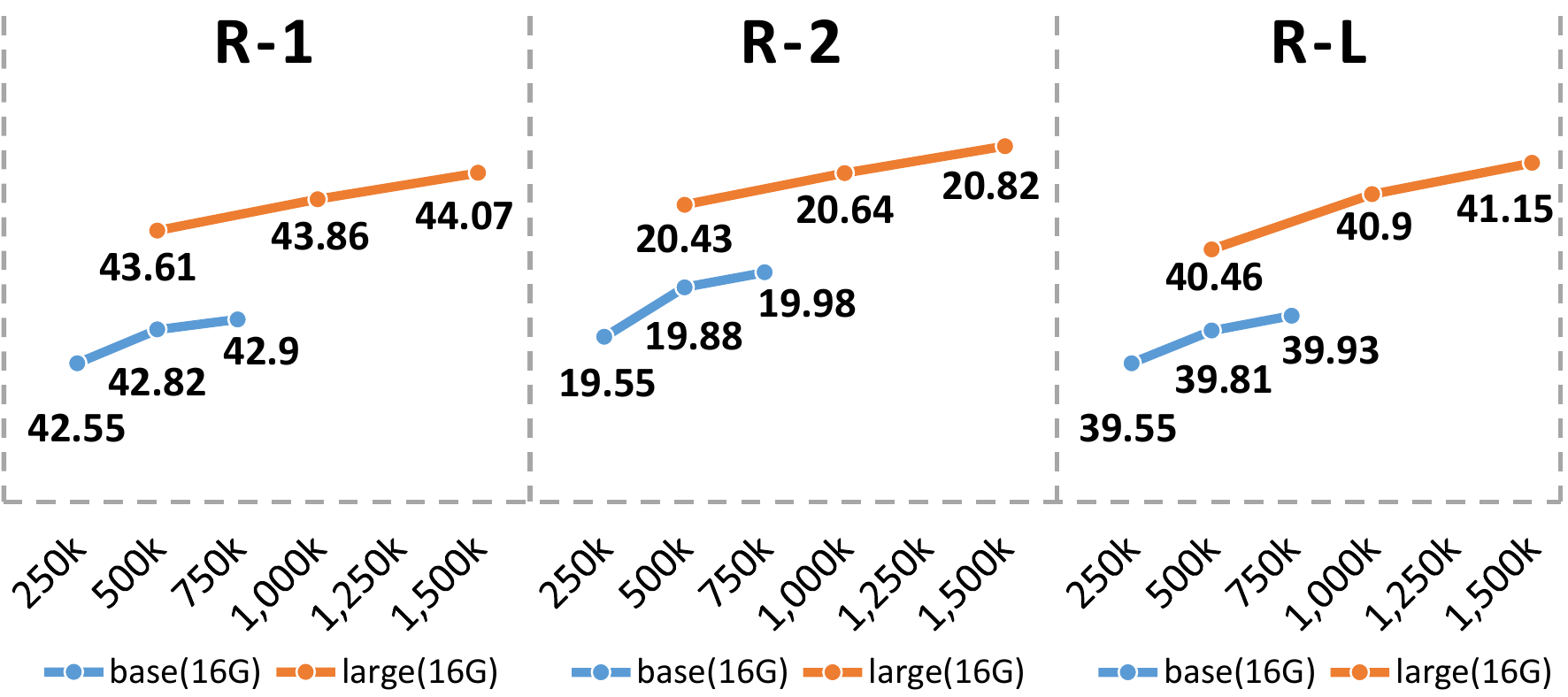}
	\caption{\label{iter_results} P$^3$LM finetuning results on CNN/DM of different pre-trained models at different iterations.}
\end{figure}

We verify that the performance of a pre-trained model improves with the increasing of training iterations within current maximum iteration number.
Figure~\ref{iter_results} shows the results of finetuned models on CNN/DM with different pre-trained models.
For both the base and large models, rouge scores increase when the models are pre-trained with more iterations.

\end{document}